\documentclass[10pt,twocolumn,letterpaper]{article}
\usepackage{wacv}              % To produce the CAMERA-READY version
\usepackage{graphicx}      % For including images
\usepackage{xcolor}
\usepackage{subcaption}    % Subfigures (use instead of deprecated subfigure)
\usepackage{amsmath, amssymb, amsfonts, mathrsfs}  % Math packages (merged into one)
\usepackage{booktabs}      % Professional tables
\usepackage{caption}       % Custom captions
\usepackage{float}         % Force figure placement
\usepackage{threeparttable} % Notes below tables
\usepackage{rotating}      % Rotating tables and figures
\usepackage{lscape}        % Landscape mode for large tables
\usepackage{breqn}         % Auto line-breaking in equations
\usepackage{pifont}        % Special symbols (e.g., checkmarks)
\usepackage{setspace}      % Adjust line spacing
\usepackage{arydshln}      % Dashed lines in tables
\usepackage{multirow}      % Multi-row table cells
\usepackage{algorithm}     % Algorithms
\usepackage{algpseudocode} % Pseudocode for algorithms
\usepackage{placeins}      % Force floats to be placed in order

\usepackage{threeparttable} % Table notes, aligning left

\usepackage{slashbox}

\usepackage[accsupp]{axessibility}  % Improves PDF readability for those with disabilities.

\usepackage[breaklinks,colorlinks]{hyperref}

\usepackage[capitalize]{cleveref}
\crefname{section}{Sec.}{Secs.}
\Crefname{section}{Section}{Sections}
\Crefname{table}{Table}{Tables}
\crefname{table}{Tab.}{Tabs.}

\def\wacvPaperID{2760} % *** Enter the WACV Paper ID here
\def\confName{CVPRW}
\def\confYear{2025}

\begin{document}

\title{Balancing Privacy and Action Performance: A Penalty-Driven Approach to Image Anonymization}

\author{Nazia Aslam$^{1,2}$, Kamal Nasrollahi$^{1,2,3}$
        \\
	$^{1}$ Aalborg University, Denmark,  $^{2}$ Pioneer Centre for AI, Denmark,
        $^{3}$ Milestone Systems, Denmark \\
	{\tt\small \{naas,kn\}@create.aau.dk}}

\maketitle

%\begin{refsection}[main]
%%%%%%%%%%%%%%---------Section---------%%%%%%%%%%%%%%
%%%%%%%%%%%%%----------Abstract----------%%%%%%%%%%%%%
\begin{abstract}
The rapid development of video surveillance systems for object detection, tracking, activity recognition, and anomaly detection has revolutionized our day-to-day lives while setting alarms for privacy concerns. It isn't easy to strike a balance between visual privacy and action recognition performance in most computer vision models. Is it possible to safeguard privacy without sacrificing performance? It poses a formidable challenge, as even minor privacy enhancements can lead to substantial performance degradation. To address this challenge, we propose a privacy-preserving image anonymization technique that optimizes the anonymizer using penalties from the utility branch, ensuring improved action recognition performance while minimally affecting privacy leakage. This approach addresses the trade-off between minimizing privacy leakage and maintaining high action performance. The proposed approach is primarily designed to align with the regulatory standards of the EU AI Act and GDPR, ensuring the protection of personally identifiable information while maintaining action performance. To the best of our knowledge, we are the first to introduce a feature-based penalty scheme that exclusively controls the action features, allowing freedom to anonymize private attributes. Extensive experiments were conducted to validate the effectiveness of the proposed method. The results demonstrate that applying a penalty to anonymizer from utility branch enhances action performance while maintaining nearly consistent privacy leakage across different penalty settings. Find code \href{https://github.com/Rabusi/Balancing-Privacy-and-Action-Performance-A-Penalty-Driven-Approach-to-Image-Anonymization}{HERE}.

\end{abstract}
%%%%%%%%%%%%%----------Abstract----------%%%%%%%%%%%%%

%%%%%%%%%%%%%%---------Section---------%%%%%%%%%%%%%%
\section{Introduction}
\label{sec:intro}

The availability of large datasets, continual improvement in learning algorithms, and advances in computing power are some pivotal reasons that make artificial intelligence (AI) an integral part of human lives, specifically in surveillance, healthcare, finance, agriculture, cybersecurity, etc. However, privacy and explainability are two significant concerns that need much attention before moving toward an ethical, fair, secure, and trustworthy AI system. In surveillance applications, human involvement significantly amplifies privacy concerns. Many global regulations, like the General Data Protection Regulation (GDPR) \cite{voigt2017eu} and the California Consumer Privacy Act (CCPA) \cite{goldman2020introduction}, have been putting forth their constant and incremental efforts to address these challenges. These laws prioritize individual privacy by imposing strict data collection and processing prerequisites.

The high computation demand to process the large video data related to human action recognition  \cite{liu2020argus,sun2022human}, elderly care monitoring systems \cite{liu2020privacy,zhang2012privacy,buzzelli2020vision} requires uploading sensitive data to a remote workstation or cloud server. This unintentional exposure reveals private attributes such as gender, race, skin color, and background objects and triggers the alarm for privacy preservation.

A simple and non-trainable approach to preserve individual privacy in action recognition systems is to downsample the video frames to a very low-resolution \cite{dai2015towards, liu2020indoor, ryoo2017privacy}. Another approach involves a bounding-box annotation technique that employs pre-trained object detectors to identify privacy-sensitive regions, which are removed or modified through synthesis \cite{ren2018learning} or blurring \cite{zhang2021multi}. However, the above two approaches struggle to achieve an optimal balance between action recognition performance and privacy leakage.

Wu et al. \cite{wu2020privacy} propose an adversarial training framework that introduces an anonymization block to remove visual privacy features from video data. This method relies on supervised privacy labels to effectively anonymize privacy-sensitive features. In contrast, Dave et al. \cite{dave2022spact} expanded the above framework to a self-supervised learning domain to eliminate time-consuming and labor-intensive private data labeling efforts. Li et al. \cite{li2023stprivacy} extend the general framework by incorporating a transformer-based anonymization block, which masks entire video tubelets that do not contribute to action recognition. However, this method requires both action and privacy labels, which can be challenging to obtain in real-world scenarios, particularly for desired privacy attributes.

Previous research mainly focuses on anonymizing the private attributes of visual data while ignoring the performance of downstream tasks such as action recognition, anomaly detection, and health monitoring system settings. In fact, in some instances, the anonymizer is so strong that it obfuscates features that lead to degradation in the performance of the utility tasks. The practical application of advanced AI systems for societal benefits becomes hazy due to compromises in accuracy, all in the name of protecting privacy. This problem arises due to the lack of awareness of the anonymizer in differentiating action and privacy-related attributes. In this work, we propose a novel approach designed to optimize the performance of downstream tasks while still ensuring individual privacy. We introduce a penalty-driven minimax optimization algorithm that selectively applies penalties to the anonymizer in order to keep action-relevant features while allowing the anonymizer to obfuscate the private attribute to the maximal extent.

In this proposed model, we incorporate a self-supervised learning strategy \cite{dave2022spact} to train the anonymizer without the need of labeled private data. This approach is particularly beneficial in real-world scenarios, where annotating private attribute data is challenging and training the model for specific private attributes would limit its applicability for cross-domain private data. Consequently, our learned anonymizer can effectively destroy the semantic information of private attributes while maintaining a better action performance. The proposed anonymization function is model-agnostic and can be applied to various action classification models \textcolor{blue}{(details in C.2 of Supp)}.

The contributions of the proposed work are summarized as follows:
\begin{enumerate}
    \item We proposed a novel penalty-driven minimax optimization algorithm that effectively anonymizes the private attributes of individuals while enhancing the performance of downstream tasks across various penalty settings.

    \item The proposed penalty regulation scheme is feature-dependent, effectively differentiate between action and privacy-sensitive features.

    \item The proposed penalty-driven anonymizer is model-agnostic and can be applied to various action classification target models.

    \item Extensive experiments have been conducted on publicly available datasets to validate the effectiveness and performance of the proposed algorithm.
\end{enumerate}

% %%%%%%%%%%%%%----------Introduction----------%%%%%%%%%%%%%
% %%%%%%%%%%%%%%%%---------Section---------%%%%%%%%%%%%%%%%

% %%%%%%%%%%%%%%%%%---------Section---------%%%%%%%%%%%%%%%%%
% %%%%%%%%%%%%%----------Related_Works----------%%%%%%%%%%%%%
\section{Related Work}
\label{sec: related work}

Recent developments in privacy-preserving vision systems are mainly categorized into 1) Downsampling-based methods, 2) Obfuscation-based methods, and 3) Learning-based methods.

Downsampling-based methods utilize low-resolution input images to anonymize the privacy-sensitive features. For example, Dai et al. \cite{dai2015towards} employed an extremely low-resolution (16 $\times$ 12) camera for video anonymization. Ryoo et al. \cite{ryoo2017privacy} demonstrate that reliable action recognition can still be achieved at extremely low resolutions by learning suitable downsampling transforms without depending on unrealistic activity-location assumptions or requiring additional specialized hardware. Furthermore, Chou et al. \cite{chou2018privacy} utilize low-resolution depth images to ensure privacy in hospital environments. Similarly, Srivastava et al. \cite{srivastav2019human} employ low-resolution images to mitigate privacy leakage in human pose estimation. Butler et al. \cite{butler2015privacy} adopt techniques like blurring and superpixel clustering to anonymize video data. Additionally, Liu et al. \cite{liu2020indoor} investigate downsampling-based approaches for privacy-preserving action recognition. Although these methods are simple and do not require privacy labels, their main limitation is to achieve an optimal balance between action recognition performance and privacy leakage.

Obfuscation-based approaches involve detecting privacy-sensitive attributes using an off-the-shelf object detector and then modifying or removing the detected regions to reduce their privacy content. Ren et al. \cite{ren2018learning} propose a solution for anonymizing faces in action detection tasks by synthesizing fake images to replace detected faces. Similarly, Zhang et al. \cite{zhang2021multi} address privacy in videos by applying semantic segmentation to identify regions of interest, followed by blurring operations to reduce privacy content. While obfuscation methods are effective in preserving privacy, they have two major limitations. First, they require domain knowledge to identify the regions of interest accurately. Second, these methods are not end-to-end and involve two steps: detecting or segmenting private objects and removing or modifying them. This significantly impacts the performance of the primary utility task.

Apart from these methods, some hardware-based solutions have been explored for protecting privacy in image-based vision systems \cite{jia2013using, pittaluga2015privacy, pittaluga2016pre}. Hinojosa et al. \cite{hinojosa2021learning, wang2019privacy} recently proposed a method for privacy-preserving human pose estimation that integrates an optical encoder at the hardware level with a software-based decoder.

Learning-based approaches involve end-to-end training to optimize the anonymizer, ensuring it effectively balances downstream task performance and privacy protection. Wu et al. \cite{wu2020privacy} proposed an adversarial learning framework that uses supervised privacy labels to obfuscate privacy features. Building on this, MaSS \cite{chen2022mass} introduced a similar framework but incorporated a compound loss to preserve specific attributes instead of completely obscuring them selectively. In contrast, Dave et al. \cite{dave2022spact} presented a self-supervised framework for privacy preservation, removing the need for labeled data for private attributes. Similarly, the self-supervised adversarial anonymization framework proposed in \cite{fioresi2023ted} for anomaly detection avoids using privacy labels and employs NT-Xent contrastive loss in the budget branch to mitigate spatial privacy leakage. In addition, Li et al. \cite{li2023stprivacy} focused on video-level anonymization by employing a video transformer to remove privacy-sensitive attributes from video data. In contrast, we focused on developing a penalty-based minimax optimization algorithm that applies penalties on the anonymizer to improve the downstream task performance while ensuring minimal privacy leakage.

% %%%%%%%%%%%%%%%%%---------Section---------%%%%%%%%%%%%%%%%%
% %%%%%%%%%%%%%----------Proposed_Method----------%%%%%%%%%%%%%
\section{Method}

The main objective of our proposed framework is to achieve a balance between the performance of the utility task and the privacy budget. Specifically, our goal is to improve the accuracy of the downstream task while ensuring individual privacy. To achieve this, we introduce a penalty-driven minimax optimization algorithm that selectively penalizes the anonymizer when it anonymizes action-related features, thus preserving their utility. At the same time, the algorithm allows the anonymizer to anonymize the private attributes of individuals to the maximum extent. 

\begin{figure*}
    \centering
    \includegraphics[width=\textwidth]{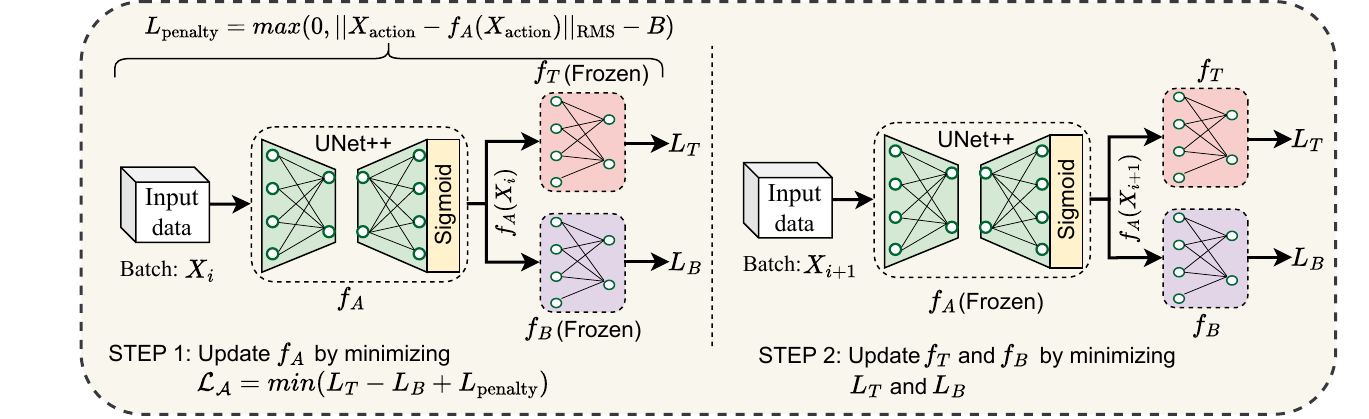}
    \caption{A penalty-driven two-step training framework for balancing action performance and privacy leakage. In Step 1, the weights of $f_A$ are updated by optimizing the gradient through $\mathcal{L_A}$ loss, while keeping the weights of $f_T$ and $f_B$ frozen. In Step 2, the weights of $f_A$ kept frozen, while the weights of $f_T$ and $f_B$ are updated using the cross-entropy loss $L_T$ and the NT-Xent contrastive loss $L_B$.}
    \label{fig:architecture}
\end{figure*}

%%%%%-------------------------------------------------------------------%%%%%

\subsection{Problem Statement}
\label{Section 3.1}

We define our problem statement similar to the previous privacy-preserving action recognition frameworks \cite{dave2022spact, wu2020privacy}, but with a different optimization objective. Consider a video \(X\), which serves as input for both an action recognition task \(f_T\) -- the utility objective,  and a private attribute classification task \(f_B\) -- the privacy budget constraint. The objective of our proposed penalty-driven optimization approach is to improve the performance of the utility task \(f_T\) while minimizing privacy leakage in the budget task \(f_B\) under different penalty settings. To achieve this, we train an anonymization function \(f_A\) to reach the optimal state \(f_A^*\) such that it will inherit critical features for the utility task while anonymizing the relevant features of the budget task. This is achieved through a penalty-driven mechanism that discourages the anonymizer from encoding privacy-relevant features while preserving action-relevant ones. This can be formulated by the following two criteria:

\textit{\textbf{Criterion 1}}: The objective of the learned anonymizer \(f_A^*\) is that it should minimally modify the utility task-related features so that the task performance remains as close to the raw data. Mathematically this can be written as:

\begin{equation}
    L_T(f_T'(f_A^*(X), Y_T)) \approx L_T(f_T'(X), Y_T).
    \label{c1}
\end{equation}

Here, \(L_T\) represents the loss function of the utility task, \(f_T'\) denotes the utility task model, and \(f_A^*(X)\) is the learned anonymization function. \(T\) refers to the utility task, and \(Y_T\) represents the labels associated with the utility task.

However, achieving Equation \ref{c1} in practice is challenging, as the anonymizer often lacks awareness of which specific features to anonymize. Some features are crucial for downstream task performance, while others are directly associated with private attributes. To address this challenge, we propose a novel penalty-driven minimax optimization approach to effectively implement Equation \ref{c1}.

\textit{\textbf{Criterion 2}}: The learned anonymizer \(f_A^*(X)\) effectively reduces privacy leakage, resulting in a significant decline in budget performance compared to the raw data. Mathematically, this can be expressed as:

\begin{equation}
    L_B(f_B'(f_A^*(X)), Y_B) \gg L_B(f_B'(X), Y_B).
    \label{c2}
\end{equation}

Here, \(L_B\) is the loss function of the budget task, \(f_B'\) is the budget task model, \(Y_B\) is the private attribute label and \(B\) denotes the budget task.

Following \cite{dave2022spact}, we adopt a self-supervised training approach for the privacy branch, eliminating the need for private attribute labels \(Y_B\) to train the anonymizer. The key idea is that increasing the self-supervised contrastive loss \(L_B\) maximizes the distance between similar features while minimizing the distance between different features for both private and action features, potentially degrading the performance of the utility branch. 

To overcome this challenge and satisfy both Criteria \ref{c1} and \ref{c2}, we propose a novel penalty-driven minimax optimization approach. The optimization objective of the anonymizer ensures that the important action features are retained for the utility branch while effectively eliminating the semantic spatial information of private attributes for the budget branch. Therefore the loss function of the anonymizer is as follows: 

\begin{align} 
    \mathcal{L}_{\text{A}} = \underset{f_A}{\arg\min}  
    &[ L_T( f_T ( f_A(X)), Y_T )  
    - L_B( f_B ( f_A(X)) )  \nonumber \\
    &+ \lambda_{\text{penalty}} \times L_{\text{penalty}}]. 
    \label{anonymizer_loss}
\end{align}

Here, $L_{\text{penalty}}$ is the penalty loss used for the anonymizer. is the penalty loss used to penalize the anonymizer specifically for the utility task. $\lambda_{\text{penalty}}$ is the weight coefficient. The penalty loss is computed by the following equation:

\begin{equation}
    L_{\text{penalty}}(X, f_A(X)) = \max \big( 0, \| (X - f_A(X)) \|_{\mathrm{RMS}} - B \big).
    \label{penalty_loss}
\end{equation}

Equation \ref{penalty_loss} ensures that the RMS loss between the input image \(X\) and the anonymized image \(f_A(X)\) does not exceed beyond \(B\). Here, \(B\) acts as a limiter and puts a constraint on the anonymizer \(f_A\), preventing excessive modification of action-relevant features to maintain the performance of the utility task. However, this penalty is applied exclusively to the action input data during the training of the anonymizer and utility task, while the anonymizer is free to maximally obfuscate private attribute features for the budget task to preserve privacy. By optimizing this objective, the anonymizer learns to preserve action-relevant features while maximally anonymizing private attributes, thus ensuring an optimal balance between action recognition performance and privacy protection.

% %%%%%-------------------------------------------------------------------%%%%%
% \vspace{-4mm}
\subsection{Proposed Approach}
\label{Section 3.2}

The proposed framework consists of three components: 1) anonymization network -- \( f_A \), 2) utility branch -- \( f_T \), and 3) budget branch -- \( f_B \). The architecture of the framework is shown in Figure \ref{fig:architecture}. An input video \(X_i = X_{\text{action}} + X_{\text{privacy}}\) is fed to \(f_A\) to generate an anonymized input \(f_A(X_i)\) that keeps the critical action features, while the private features are removed. The optimization of \( f_A \) involves three losses: \( L_T \) (cross-entropy loss) to minimize classification error in \( f_T \), \( L_B \) (NT-Xent contrastive loss which increases the disagreement between the same frames while increasing the agreement between the different frame) in \( f_B \) to prevent privacy leakage by maximizing it, and \[L_{\text{penalty}} = \max(0, \|X_{\text{action}} - f_T(f_A(X_{\text{action}}))\|_{\text{RMS}} - B)\] which prevents excessive degradation of action features. Here, \( B \) acts as a threshold that limits the anonymization of action features. No penalty is applied on \( X_{\text{privacy}} \) to ensure complete anonymization of private attributes. A two-step training approach is carried out to optimize \( f_A \), incorporating self-supervised contrastive learning for the budget branch and penalty-driven optimization for the anonymizer. In Step 1, the weights of \( f_A \) are updated by optimizing Equation \ref{anonymizer_loss} while keeping \( f_T \) and \( f_B \) frozen. While in Step 2, the weights of \( f_A \) are kept frozen, and \( f_T \), \( f_B \) are updated by minimizing \( L_T \) and \( L_B \) respectively. This adversarial learning dynamic ensures that in Step 1, \( f_A \) learns to remove private features, while in Step 2, \( f_B \) learns to reconstruct them, simulating an inference attack. Over time, this competition strengthens the anonymizer, making it highly effective against privacy threats while preserving action recognition performance.

% %%%%%-------------------------------------------------------------------%%%%%

\subsubsection{Anonymization Framework}
\label{Section 3.2.1}

The anonymization function \(f_A\) is a trainable function that transforms the input data in such a way that it keeps the essential features for the utility task while minimizing the relevant features for the budget task. To achieve this, we employ an encoder-decoder neural network, which is pretrained to initialize as an identity function. The pretraining of \(f_A\) is performed through input reconstruction using the \(L_1\) loss, formulated as follows: 

\begin{equation}
    \mathcal{L}_{L1} = \sum_{c=1}^{C} \sum_{h=1}^{H} \sum_{w=1}^{W} |X_{c,h,w} - \hat{X}_{c,h,w}|.
    \label{l1 loss}
\end{equation}

Here, \(X\) represents the input video data, while \(\hat{X}\) denotes its reconstructed version. The parameters \(C\), \(H\), and \(W\) correspond to the channel, height, and width of the input data, respectively.

\subsubsection{Penalty Driven Anonymization Training}
\label{Section 3.2.2}

Figure \ref{fig:architecture} illustrates the basic architecture of the proposed framework, designed to implement the formulation in Equation \ref{anonymizer_loss}. The model takes input \(X\) and processes it through an anonymization function \(f_A\), which applies a transformation to generate the anonymized video \(f_A(X)\). During training, this anonymized output is simultaneously fed into two different modules: the utility task module \(f_T\), responsible for learning the target action recognition task, and the privacy prediction module \(f_B\), which assesses privacy leakage. All three modules, \(f_A\), \(f_T\), and \(f_B\), are learnable deep neural networks and are trained in an end-to-end manner. 

The anonymizer $f_A$ inherently transforms all incoming input data, leading to the suppression of both privacy-sensitive attributes and essential features required for action recognition. This degradation negatively impacts the performance of the downstream utility task. To mitigate this issue, we adopt a penalty-driven approach that guides the anonymizer. This strategy ensures that the anonymizer maximally anonymize features relevant to the budget branch while preserving crucial information necessary for the utility branch, thereby maintaining task performance. The training process involves the optimization of the loss function $\mathcal{L}_A$, which combines the utility loss $L_T$, budget loss $L_B$ and penalty loss $L_{\text{penalty}}$. Following \cite{dave2022spact}, we have opted for the self-supervised contrastive learning framework for the budget loss, which tries to maximize the distance between the same features of the input data while minimizing the distance between the different features of the input data. This strategy is opted just as opposite to the original SimCLR\cite{chen2020simple} work because here we want to degrade the learning process of the anonymizer for the budget task so that it will not be able to learn the essential features for the private data and anonymize the private attributes. However, we have opted for the cross entropy loss for the utility branch for the multi-class classification of the action categories. In addition, we have implied another loss function $L_{\text{penalty}}$, which is an RMS loss between the input data $X$ and the anonymized data $f_A(X)$ with a penalty term $B$ as elaborated in Equation \ref{penalty_loss}. 

Equation \ref{penalty_loss} indicates that if the degradation of the input data $X$ exceeds the limit $B$, a penalty is applied on the anonymizer's loss function. This prevents the anonymizer from anonymizing beyond $B$, ensuring the performance of the utility task is not degraded. It is important to note that Equation \ref{penalty_loss} is applied exclusively to action-related features during the training of the anonymizer and utility task model. The optimization of this hybrid loss function $\mathcal{L_A}$, combining the utility loss $L_T$, privacy loss $L_B$, and penalty loss $L_{\text{penalty}}$ enables the anonymizer $f_A$ to find an optimal transformation that retains discriminative features for the utility task while suppressing privacy-sensitive attributes for the budget task. 

This end-to-end optimization ensures that the anonymized data is beneficial for utility performance while reducing privacy leakage. Once trained, the anonymization function can be deployed on edge devices (e.g., cameras) to process video data before transmission to a backend system (e.g., cloud) for task-specific analysis. The proposed framework offers a flexible and adaptive solution for privacy-preserving visual recognition.

% %%%%%-------------------------------------------------------------------%%%%%

\subsubsection{Minimax Optimization:} 
\label{Section 3.3.3}
Following the approach of \cite{dave2022spact, wu2020privacy}, we adopt a minimax optimization strategy to optimize the loss function of \(f_A\), represented by Equation \ref{anonymizer_loss}. The goal is to minimize the utility task loss \(L_T\) and the penalty loss \(L_{\text{penalty}}\) while maximizing the budget task loss \(L_B\). This ensures improved action performance with reduced privacy leakage. The complete training algorithm of the anonymizer, utility, and budget module is explained in Algorithm \ref{algorithm}.

% %%%%%-------------------------------------------------------------------%%%%%
\begin{algorithm}[]
\caption{Penalty driven framework}
\label{algorithm}
\begin{algorithmic}[1]

\State \textbf{Inputs:}
\State Datasets: \(X_{\text{input}}\)
\State Epochs: \(e_{\text{anon}}, e_{\text{action}}, e_{\text{privacy}}\)
\State Learning Rates: \(\alpha_A, \alpha_B, \alpha_T\)
\State Hyperparameters: \(\lambda_P\)

\State \textbf{Output:} \(\theta_A, \theta_T', \theta_B'\)

\State \textbf{Learnable Parameter}: \(\theta_A, \theta_T, \theta_B\)

\State \textbf{Initialization:}
\State \(\theta_A \gets \theta_A - \alpha_A \nabla_{\theta_A} \mathcal{L}_{L1}(\theta_A)\) 
\State \Comment{Equation \ref{l1 loss}}
\State \(\theta_B\): SimCLR \cite{chen2020simple} weights trained on ImageNet \cite{deng2009imagenet} 
\State \(\theta_T\): pre-trained weight on Kinetics400 \cite{carreira2017quo}

\State \textbf{Anonymization Training:}
\For{\(e_0 \gets 1\) to \(e_{\text{anon}}\)}
    \State \textbf{Step-1:} update anonymizer parameters 
    \State     
    \(\begin{aligned}
        &\theta_A \leftarrow \theta_A - \alpha_A \nabla_{\theta_A} [ L_T (\theta_A, \theta_T) - L_B (\theta_A, \theta_B) \\
        &+ \lambda_P L_P (\theta_A, \theta_T)]
    \end{aligned}\)
    \State \textbf{Step-2:} update utility and budget parameters
    \State \(\theta_T \gets \theta_T - \alpha_T \nabla_{\theta_T} L_T(\theta_T, \theta_A)\)
    \State \(\theta_B \gets \theta_B - \alpha_B \nabla_{\theta_B} L_B(\theta_B, \theta_A)\)
\EndFor

\State \textbf{Privacy-Preserved Action Recognition Training:}
\For{\(e_0 \gets 1\) to \(e_{\text{action}}\)}
    \State \(\theta_T' \gets \theta_T' - \alpha_T' \nabla_{\theta_T'} L_{T}(\theta_T', \theta_A^*)\)
\EndFor

\State \textbf{Privacy Leakage Training:}
\For{\(e_0 \gets 1\) to \(e_{\text{privacy}}\)}
    \State \(\theta_B' \gets \theta_B' - \alpha_B' \nabla_{\theta_B'} L_{T}(\theta_B', \theta_A^*)\)
\EndFor

\end{algorithmic}
\end{algorithm}

%%%%%%%%%%%%%%%%%%%%%%--------------Experiments--------------%%%%%%%%%%%%%%%%%%%%%%
\section{Experiments}

\subsection{Datasets:}
\label{4.1}

\textbf{UCF101} \cite{soomro2012ucf101} and \textbf{HMDB51} \cite{kuehne2011hmdb} are the two most commonly used publically available action datasets with 101 and 51 day-to-day actions.

The \textbf{PAHMDB} \cite{wu2020privacy} is a subset of the HMDB51 dataset, composed of 515 video data with 51 video action labels. Along with the video action label, this dataset has annotated frames for the five different private attributes like skin color, race, gender, nudity, and relationship.

The \textbf{VISPR} \cite{orekondy2017towards} dataset contains 22,167 different images labeled with 68 private attributes. It is a multi-class classification benchmark dataset for evaluating privacy-preserving techniques. We utilize two subsets of the VISPR dataset, named VISPR1 and VISPR2. \textcolor{blue}{(Additional details are in Section A of Supp.)}

The \textbf{VPUCF} \cite{li2023stprivacy} and \textbf{VPHMDB} \cite{li2023stprivacy} is a large video-annotated private dataset. The VPHMDB is created using the HMDB51 dataset, which contains 51 different actions from 6,849 videos, while the VPUCF is created using the UCF101 dataset, which has 101 action classes from 13,320 videos. Both datasets are annotated at the video level with private attributes, making them valuable resources for privacy-related research. The annotations include five private attributes: face, skin color, gender, nudity, and familiar relationships.

% %%%%%%%--------------------------------------------------%%%%%%%%

\subsection{Implementation details:}
\label{4.2}

\textbf{Network Architecture:} The anonymizer \(f_A\) employs a UNet++ \cite{zhou2018unet++} architecture, functioning as an encoder-decoder structure. It processes input data of size \(224 \times 224\) and outputs an anonymized version with the same dimensions. I3D ResNet-50 \cite{carreira2017quo} is used as the action classifier \(f_T\), and a standard ResNet-50 \cite{he2016deep} is used for the private attribute classifier \(f_B\). 

%%%%%%%--------------------------------------------------%%%%%%%%
\noindent
\textbf{Initialization and Training:} For the initialization of \(f_A\), we trained UNet++ for 100 epochs using \(L_1\) reconstruction loss (Eq. \ref{l1 loss}). \(f_T\) is initialized with pretrained weights from the Kinetics400 action dataset \cite{carreira2017quo}, while \(f_B\) was initialized using self-supervised learning (SSL) weights from SimCLR \cite{chen2020simple} trained on ImageNet \cite{deng2009imagenet}. After initialization, the anonymizer was trained for 100 epochs using the Adam optimizer \cite{kingma2014adam}. A batch size of 8 was chosen to train the anonymizer. For the self-supervised learning (SSL) of \(f_B\), a margin of \(\mu\) = 1 was used as the default setting. To evaluate \(f_T\), a batch size of 16 is used for 100 epochs, and for \(f_B\), we used a batch size of 32 and a base learning rate of 1e-3, following a linear warmup and a step-based scheduler that reduced the learning rate by a factor of 1/5 upon loss stagnation. All experiments were conducted using the PyTorch framework on NVIDIA TESLA A100 GPUs.

% %%%%%%%--------------------------------------------------%%%%%%%%
\noindent
\textbf{Input details:} For all experiments, we first crop each image to 0.8 of its original scale and then resize it to an input resolution of \(224 \times 224\). Each clip consists of 16 frames, sampled from a random starting point with a skip rate of 4. During anonymization training, we apply standard augmentations, including random erase, random crop, horizontal flip, and random color jitter.

% %%%%%%%--------------------------------------------------%%%%%%%%

\subsection{Training and Evaluation Protocols:}
\label{4.3}

To evaluate the performance of the learned anonymizer \(f_A^*\), we follow the standard cross-dataset training and evaluation protocols \cite{dave2022spact,wu2020privacy,wu2018towards}. This cross-dataset training and evaluation approach involves jointly utilizing an action dataset and a private attribute dataset to train the anonymizer. The objective is to ensure that the anonymized output effectively learns action recognition (evaluated on an action dataset) while simultaneously suppressing the prediction of multiple privacy attributes (evaluated on private data). To evaluate the action performance on the anonymized dataset \(f_A^*(X)\) on the target model \(f_T'\), we utilize Top-1 accuracy \((A_T^1)\). Additionally, the privacy leakage of the target privacy model \(f_B'\) on the test set of the private attribute dataset is assessed using class-wise mean average precision (cMAP) and the \(F_1\) score.

% %%%%%%%-------------------PUSHUP and BRUSHHAIR----------------------%%%%%%%%

\begin{figure*}[ht]
	\centering
	\begin{subfigure}[t]{0.5\linewidth}
		\centering
		\includegraphics[width=8cm,height=6cm]{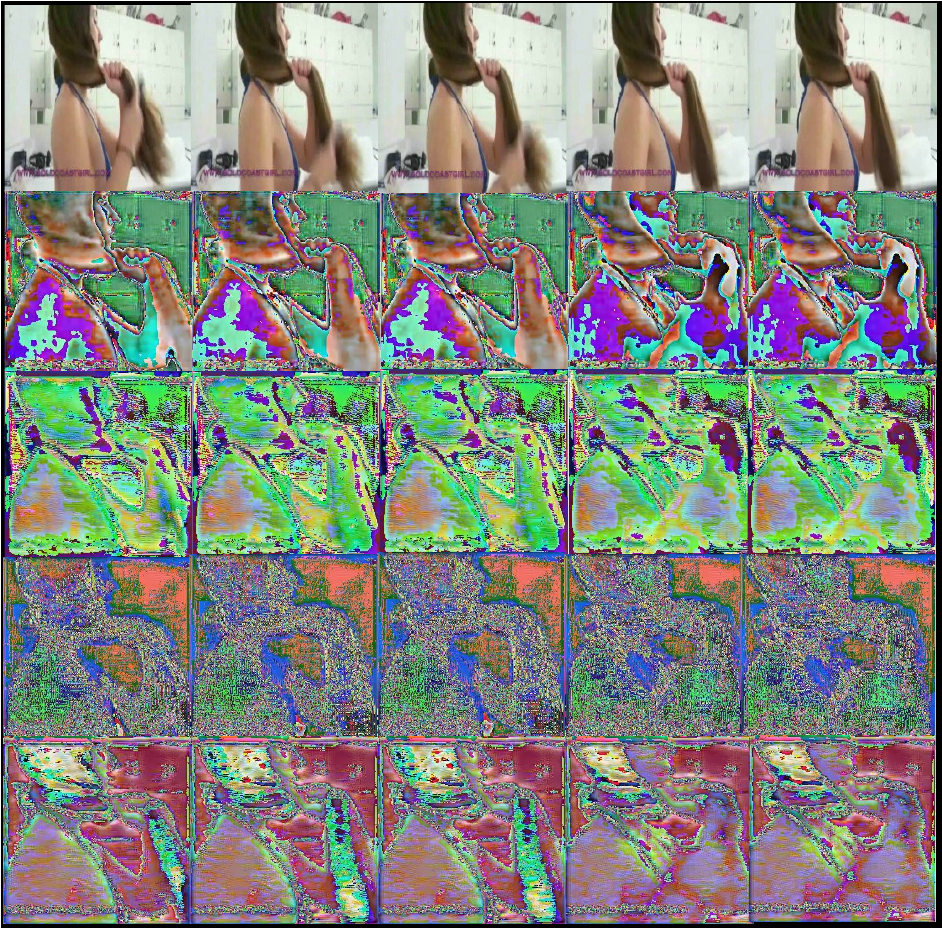}
		\caption{Brush Hair Action}
		\label{fig:brushhair}
	\end{subfigure}%
	\begin{subfigure}[t]{0.5\linewidth}
		\centering
		\includegraphics[width=8cm,height=6cm]{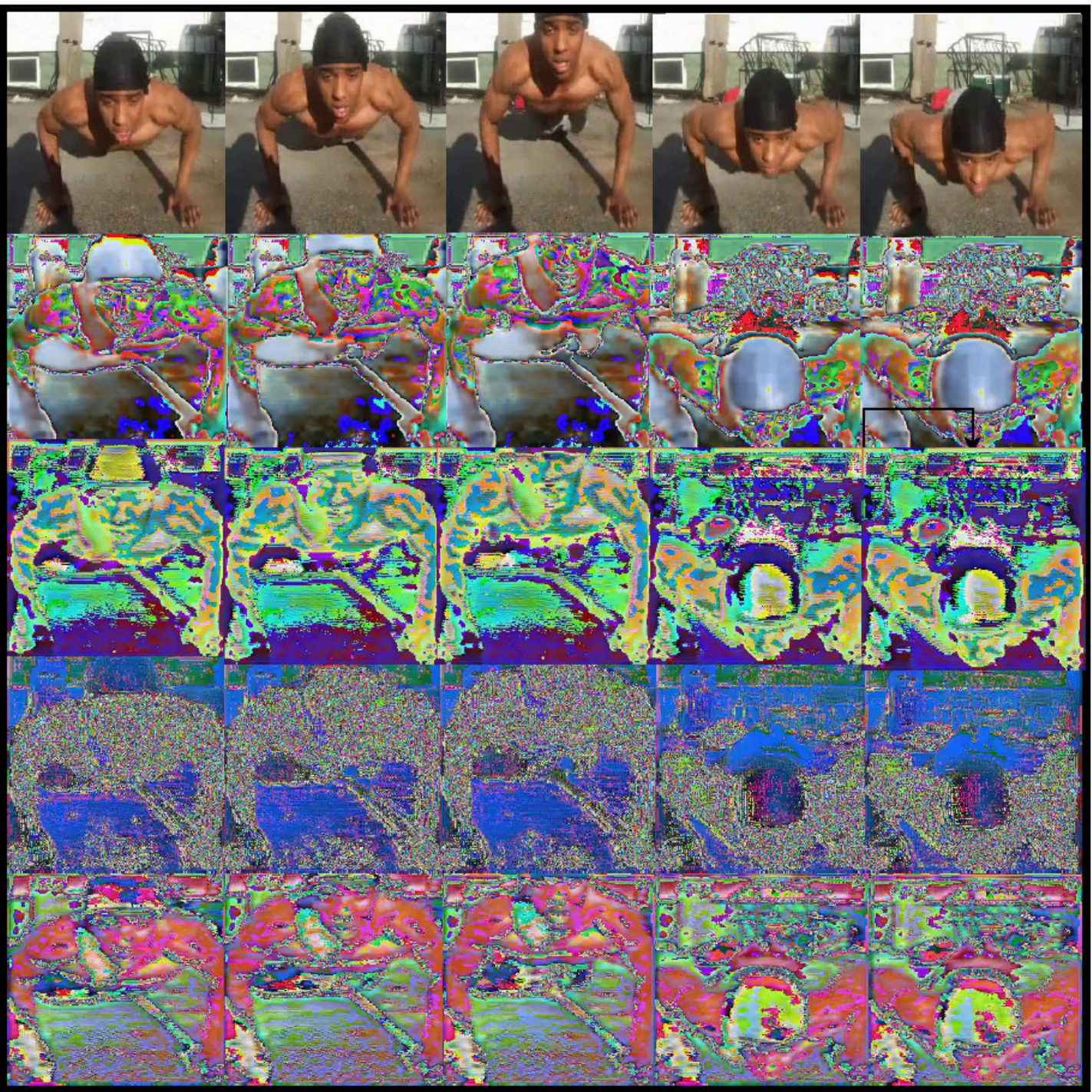}
		\caption{Push-up Action}
		\label{fig:pushup}
	\end{subfigure}
	\caption{Anonymized frames of two different actions from the HMDB51 dataset across different penalty settings. Top to bottom: raw image, followed by $B = 0.3$, $B = 0.5$, $B = 0.7$, and $B = 0.9$.}
	\label{fig:actions}
\end{figure*}

% %%%%%%%---------------------TABLE-----------------------------%%%%%%%%

\begin{table*}[ht]
\centering
\resizebox{\textwidth}{!}{
\begin{tabular}{cl| ccc| ccc| ccc}
\hline
\multicolumn{2}{c|}{\multirow{3}{*}{Method}} & \textbf{UCF101} & \multicolumn{2}{c|}{\textbf{VISPR1}} & \multicolumn{3}{c|}{\textbf{PA-HMDB}} & \multicolumn{3}{c}{\textbf{VPHMDB}} 
\\ 
\multicolumn{2}{c|}{} & Action & \multicolumn{2}{c|}{Privacy} & \multicolumn{1}{c}{Action} & \multicolumn{2}{c|}{Privacy} & \multicolumn{1}{c}{Action} & \multicolumn{2}{c}{Privacy} 
\\ 
\multicolumn{2}{c|}{} & Top-1 ($\uparrow$ \%) & cMAP ($\downarrow$ \%) & F1 ($\downarrow$ \%) & \multicolumn{1}{c}{Top-1 ($\uparrow$ \%)} & \multicolumn{1}{c}{cMAP ($\downarrow$ \%)} & F1 ($\downarrow$ \%) & \multicolumn{1}{c}{Top-1 ($\uparrow$ \%)} & \multicolumn{1}{c}{cMAP ($\downarrow$ \%)} & F1 ($\downarrow$ \%)
\\ \hline
\multicolumn{2}{c|}{Raw data} & 78.88 & 64.41 & 0.555 & 65.24 & 70.2 & 0.396 & 81.56 & 76.62 & 0.684 \\ 
\multicolumn{2}{c|}{Downsample-2×} & 54.11 & 57.23 & 0.483 & 36.1 & 61.2 & 0.111 & 40.80 & 0.601 & 71.35 \\ 
\multicolumn{2}{c|}{Downsample-4×} & 39.65 & 50.07 & 0.379 & 25.8 & 41.4 & 0.081 & 31.32 & 69.79 & 0.594 \\ 
\multicolumn{2}{c|}{Obf-Blackening} & 53.13 & 56.39 & 0.457 & 34.2 & 63.8 & 0.386 & 38.27 & 74.06 & 0.649 \\ 
\multicolumn{2}{c|}{Obf-StrongBlur} & 55.59 & 55.94 & 0.456 & 36.4 & 64.4 & 0.243 & 40.91 & 74.33 & 0.655 \\ 
\multicolumn{2}{c|}{Obf-WeakBlur} & 61.52 & 63.52 & 0.523 & 41.5 & 69.1 & 0.384 & 47.24 & 75.11 & 0.663 \\ 
\hline
\multicolumn{2}{c|}{VITA \cite{wu2020privacy}} & 62.10 & \textbf{55.32} & 0.461 & 42.3 & \textbf{62.3} & 0.194 & 48.11 & 73.89 & 0.638 \\ 
\multicolumn{2}{c|}{SPACT \cite{dave2022spact}} & 62.03 & 57.43 & 0.473 & 43.1 & 62.7 & 0.176 & -- & -- & -- \\ 
\multicolumn{2}{c|}{STPrivacy \cite{li2023stprivacy}} & -- & -- & -- & 50.61 & 68.76 & 0.523 & 50.73 & 72.48 & 0.613 \\ 
\hline
\multirow{4}{*}{Ours} & B = 0.3 & \textbf{78.26} (\textcolor{cyan}{$\downarrow$ 0.26}) & 57.56 (\textcolor{cyan}{$\downarrow$ 6.48}) & 0.468 (\textcolor{cyan}{$\downarrow$ 0.074})& \textbf{60.58} (\textcolor{cyan}{$\downarrow$ 4.66}) & 65.67 (\textcolor{cyan}{$\downarrow$ 4.53}) & 0.253 (\textcolor{cyan}{$\downarrow$ 0.143}) & \textbf{74.61} (\textcolor{cyan}{$\downarrow$ 6.95}) & 70.01 (\textcolor{cyan}{$\downarrow$ 6.61}) & 0.514 (\textcolor{cyan}{$\downarrow$ 0.170}) \\ 

& B = 0.5 & 75.57 (\textcolor{cyan}{$\downarrow$ 3.31}) & 58.04 (\textcolor{cyan}{$\downarrow$ 6.01}) & 0.493 (\textcolor{cyan}{$\downarrow$ 0.049}) & 55.53 (\textcolor{cyan}{$\downarrow$ 9.71}) & 65.02 (\textcolor{cyan}{$\downarrow$ 5.18}) & 0.191 (\textcolor{cyan}{$\downarrow$ 0.205}) & 73.79 (\textcolor{cyan}{$\downarrow$ 7.77}) & 70.75 (\textcolor{cyan}{$\downarrow$ 5.87}) & \textbf{0.519} (\textcolor{cyan}{$\downarrow$ 0.165}) \\ 
 
& B = 0.7 & 72.99 (\textcolor{cyan}{$\downarrow$ 5.89}) & \textbf{57.10} (\textcolor{cyan}{$\downarrow$ 6.94}) & \textbf{0.450} (\textcolor{cyan}{$\downarrow$ 0.092}) & 55.14 (\textcolor{cyan}{$\downarrow$ 10.1}) & 65.19 (\textcolor{cyan}{$\downarrow$ 5.01}) & 0.124 (\textcolor{cyan}{$\downarrow$ 0.272}) & 69.24 (\textcolor{cyan}{$\downarrow$ 12.32}) & 70.56 (\textcolor{cyan}{$\downarrow$ 6.06}) & 0.516 (\textcolor{cyan}{$\downarrow$ 0.168}) \\ 
 
& B = 0.9 & 71.89 (\textcolor{cyan}{$\downarrow$ 6.99}) & 57.17 (\textcolor{cyan}{$\downarrow$ 6.88}) & 0.460 (\textcolor{cyan}{$\downarrow$ 0.082}) & 48.54 (\textcolor{cyan}{$\downarrow$ 16.7}) & 65.88 (\textcolor{cyan}{$\downarrow$ 4.32}) & 0.173 (\textcolor{cyan}{$\downarrow$ 0.173}) & 65.63 (\textcolor{cyan}{$\downarrow$ 19.93}) & \textbf{69.95} (\textcolor{cyan}{$\downarrow$ 6.67}) & 0.511 (\textcolor{cyan}{$\downarrow$ 0.173}) \\ 
 \hline
\end{tabular}
}

\caption{Comparison of different privacy-preserving methods on the \textbf{known action} and \textbf{private attributes} with our penalty settings. $\downarrow$\% is the relative drop from the raw data, and -- indicates the model does not perform the experiment on the dataset. High performance in action and low performance in privacy are considered as better.}
\label{known data}
\vspace{-4mm}
\end{table*}

% %%%%%%-----------------TABLE-----------------%%%%%%%%

\begin{table*}[ht]
\centering
\resizebox{0.9\textwidth}{!}{
\begin{tabular}{cc|cccc|ccc}
\hline
\multicolumn{2}{c|}{\multirow{3}{*}{Method}} & \textbf{UCF101} $\rightarrow$ \textbf{HMDB51} & \textbf{UCF101} $\rightarrow$ \textbf{PAHMDB} & \multicolumn{2}{c|}{\textbf{VISPR1} $\rightarrow$ \textbf{VISPR2}} & \multicolumn{3}{c}{\textbf{VPHMDB} $\rightarrow$ \textbf{VPUCF}} \\
\multicolumn{2}{c|}{} & Action & Action & \multicolumn{2}{c|}{Privacy} & Action & \multicolumn{2}{c}{Privacy} \\
\multicolumn{2}{c|}{} & Top-1 ($\uparrow$ \%) & Top-1 ($\uparrow$ \%) & cMAP ($\downarrow$ \%) & F1 ($\downarrow$ \%) & Top-1 ($\uparrow$ \%) & cMAP ($\downarrow$ \%) & F1 ($\downarrow$ \%) \\
\hline
\multicolumn{2}{c|}{Raw data} & 47.51 & 67.18 & 57.63 & 0.434 & 92.23 & 76.62 & 0.699 \\ 
\hline
\multicolumn{2}{c|}{VITA \cite{wu2020privacy}} & 33.2 & 40.6 & 49.6 & 0.399 & 77.48 & 76.02 & 0.669 \\ 
\multicolumn{2}{c|}{SPACT \cite{dave2022spact}} & 34.1 & 42.8 & \textbf{47.1} & 0.386 & 78.13 & 75.98 & 0.661 \\ 
\multicolumn{2}{c|}{STPrivacy \cite{li2023stprivacy}} & -- & -- & -- & -- & 81.04 & 74.60 & 0.645 \\ 
\hline
\multirow{4}{*}{\textbf{Ours}} & B = 0.3 & \textbf{41.17} (\textcolor{cyan}{$\downarrow$ 6.34}) & \textbf{62.91} (\textcolor{cyan}{$\downarrow$ 4.27}) & 50.22 (\textcolor{cyan}{$\downarrow$ 7.41}) & 0.391 (\textcolor{cyan}{$\downarrow$ 0.043}) & \textbf{91.91} (\textcolor{cyan}{$\downarrow$ 0.32}) & \textbf{70.91} (\textcolor{cyan}{$\downarrow$ 8.71})  & \textbf{0.612} (\textcolor{cyan}{$\downarrow$ 0.087})\\  
& B = 0.5 & 40.71 (\textcolor{cyan}{$\downarrow$ 6.8}) & 58.25 (\textcolor{cyan}{$\downarrow$ 8.93}) & 49.5 (\textcolor{cyan}{$\downarrow$ 8.13}) & 0.369 (\textcolor{cyan}{$\downarrow$ 0.065}) & 87.58 (\textcolor{cyan}{$\downarrow$ 4.65}) & 70.95 (\textcolor{cyan}{$\downarrow$ 8.67}) & 0.613 (\textcolor{cyan}{$\downarrow$ 0.086}) \\  
& B = 0.7 & 40.01 (\textcolor{cyan}{$\downarrow$ 7.5}) & 57.82 (\textcolor{cyan}{$\downarrow$ 9.36}) & 49.92 (\textcolor{cyan}{$\downarrow$ 7.71}) & 0.371 (\textcolor{cyan}{$\downarrow$ 0.063}) & 89.17 (\textcolor{cyan}{$\downarrow$ 3.06}) & 70.93 (\textcolor{cyan}{$\downarrow$ 8.69}) & 0.612 (\textcolor{cyan}{$\downarrow$ 0.087}) \\  
& B = 0.9 & 39.1 (\textcolor{cyan}{$\downarrow$ 8.4}) & 54.95 (\textcolor{cyan}{$\downarrow$ 12.23}) & 50.10 (\textcolor{cyan}{$\downarrow$ 7.53}) & \textbf{0.352} (\textcolor{cyan}{$\downarrow$ 0.082}) & 78.95 (\textcolor{cyan}{$\downarrow$ 13.28}) & 70.98 (\textcolor{cyan}{$\downarrow$ 8.64}) & 0.614 (\textcolor{cyan}{$\downarrow$ 0.085}) \\  
\hline
\end{tabular}
}
\caption{Comparison of different privacy-preserving methods on \textbf{novel actions} and \textbf{novel private attributes} datasets. $\downarrow$\% represents the relative drop from raw data, and -- indicates the model does not perform the experiment on the dataset. High performance in action and low performance of privacy is considered as better.}
\label{unseen_data}

\end{table*}

% %%%%%%%--------------------------------------------------%%%%%%%%

\subsubsection{Evaluation of anonymizer on known action and private attributes}
\label{4.3.1}

In this setting of the experiment, \(f_A\) and \(f_T\) was trained on \((X_{\text{action}}^t, Y_T^t)\) to learn the action features and \((X_{\text{privacy}}^t, Y_B^t)\) to train the \(f_A\) and \(f_B\) to suppress the private attributes of the data. However, note that the 
two pipelines share the same parameters for \(f_A\). Also, we have followed the SSL framework from \cite{dave2022spact}, therefore, we have not used \(Y_B^t\) to train the anonymizer. After the cross dataset training a learned anonymizer \(f_A^*\) weight has been obtained, and the target task model \(f_T\) and budget task \(f_B\) model has been discarded from the model. To evaluate the action performance a new classifier \(f_T'\) has been trained on action annotated dataset \(f_T'(f_A^*((X_{\text{action}}, Y_T^t))\) and tested on \(f_T'(f_A^*(X_{\text{action}}^e, Y_T^e))\). For the privacy leakage a new private attribute classifier \(f_B'\) has been trained on \(f_B'(f_A^*(X_{\text{privacy}}^t, Y_B^t))\) and evaluated on the \(f_B'(f_A^*(X_{\text{privacy}}^e, Y_B^e))\). 

% %%%%%%%--------------------------------------------------%%%%%%%%
\noindent
\\
\textbf{UCF101-VISPR1 cross dataset training and evaluation:} In this setting of the experiment \(X_{\text{action}}^t\) = UCF101 trainset (split-1) and \(X_{\text{privacy}}^t\) = VISPR1 trainset, and  \(X_{\text{action}}^e\) = UCF101 testset (split-1) and \(X_{\text{privacy}}^e\) = VISPR1 testset.
% %%%%%%%--------------------------------------------------%%%%%%%%
\\
\noindent
\\
\textbf{HMDB51-VISPR1 cross dataset training and same dataset evaluation:} In this setting of the experiment \(X_{\text{action}}^t\) = HMDB51 trainset and \(X_{\text{privacy}}^t\) = VISPR1 trainset, and  \(X_{\text{action}}^e\) = PA-HMDB and \(X_{\text{privacy}}^e\) = PA-HMDB. 
% %%%%%%%--------------------------------------------------%%%%%%%%
\\
\noindent
\\
\textbf{VPHMDB-VISPR1 cross dataset training and same dataset evaluation:} In this setting of the experiment \(X_{\text{action}}^t\) = VPHMDB trainset and \(X_{\text{privacy}}^t\) = VISPR1 trainset, and  \(X_{\text{action}}^e\) = VPHMDB testset and \(X_{\text{privacy}}^e\) = VPHMDB testset.
\\
\noindent
% %%%%%%%--------------------------------------------------%%%%%%%% 

The comparative analysis of the proposed framework with the baselines is shown in Table \ref{known data}. We have conducted experiments on our proposed penalty-driven approach using four different values of \(B\) = 0.3, 0.5, 0.7, and 0.9. As shown in Table \ref{known data}, the action performance at \(B\) = 0.3 is closest to the raw data for UCF101, while for other dataset \(B\) = 0.3 reflects the highest action performance among all \(B\)'s settings. Increasing the penalty from \(B\) = 0.5 to \(B\) = 0.9 leads to a decline in action performance. This indicates that the model is anonymizing the action features more. However, we observed that the privacy leakage remains nearly constant across or little bit compromised with a different values of \(B\).
Compared to \cite{dave2022spact}, our model exhibits similar privacy leakage, whereas \cite{wu2020privacy} achieves approximately 1.5\% lower privacy leakage on the VISPR1 dataset. Despite this, our model outperforms in action classification, showing a 16.10\% improvement on the UCF101 dataset. For the PA-HMDB dataset, our approach improves action performance by 18.23\%, with a compromised privacy leakage of 3.37\%. For VPHMDB dataset, our approach performs better for action performance and lowers the privacy leakage than the \cite{wu2020privacy,li2023stprivacy}. 
Notably, at \(B\) = 0.3, we achieve a substantial increase in action performance while maintaining privacy leakage. This suggests that selecting \(B\) = 0.3 provides a better balance, ensuring strong downstream task performance without compromising privacy protection. The anonymized results of our methods is shown in Figure \ref{fig:pushup} and \ref{fig:brushhair} with different penalty settings.

% %%%%%%%--------------------------------------------------%%%%%%%%

\subsubsection{Evaluation of anonymizer on novel action and private attributes}
\label{4.3.2}

It is also important to evaluate the performance of the trained anonymizer \(f_A^*\) on the unseen dataset for the generalization of the model. We trained the anonymizer on one dataset and evaluated it on a different dataset to ensure that the model had not encountered the test data during training. Table \ref{unseen_data} provides a comparative analysis of the proposed framework against baseline (learning-based approaches), showing that our penalty-driven method achieves superior performance on novel data. For $B = 0.3$, the action performance remains close to that of raw data, while privacy leakage is reduced by 3.69\% compared to \cite{li2023stprivacy} and remains stable across all $B$'s settings. This indicates that applying a penalty on the anonymizer for action features helps preserve action performance while allowing maximum obfuscation of private attributes, hence a significant drop in cMAP and F1 scores has been found compared to raw data.
% %%%%%%%--------------------------------------------------%%%%%%%%
\\
\noindent
\\
\textbf{UCF101-HMDB51 and UCF101-PAHMDB novel dataset training and evaluation:} In this setting of the experiment \(X_{\text{action}}^t\) = UCF101 trainset (split-1), \(X_{\text{action}}^{\text{nt}}\) = HMDB51 trainset and \(X_{\text{action}}^{\text{ne}}\) = HMDB51 testset/PAHMB and for \(X_{\text{privacy}}^t\) = VISPR1 trainset, \(X_{\text{privacy}}^{\text{nt}}\) = VISPR2 trainset and \(X_{\text{privacy}}^{\text{ne}}\) = VISPR2 testset.
%%%%%%%--------------------------------------------------%%%%%%%%
\\
\noindent
\\
\textbf{VPHMDB-VPUCF novel dataset training and evaluation:} In this setting \(X_{\text{action}}^t\) = VPHMDB trainset, \(X_{\text{action}}^{\text{nt}}\) = VPUCF trainset and \(X_{\text{action}}^{\text{ne}}\) = VPUCF testset and \(X_{\text{privacy}}^t\) = VISPR1 trainset, \(X_{\text{privacy}}^{\text{nt}}\) = VPHMDB trainset and \(X_{\text{privacy}}^{\text{ne}}\) = VPHMDB testset.

% %%%%%%%--------------------------------------------------%%%%%%%%

\subsection{Ablation Analysis:}
\label{4.4}

\subsubsection{Effect of different values of $B$}

\begin{figure}[ht]
    \centering
    \includegraphics[width=0.95\linewidth]{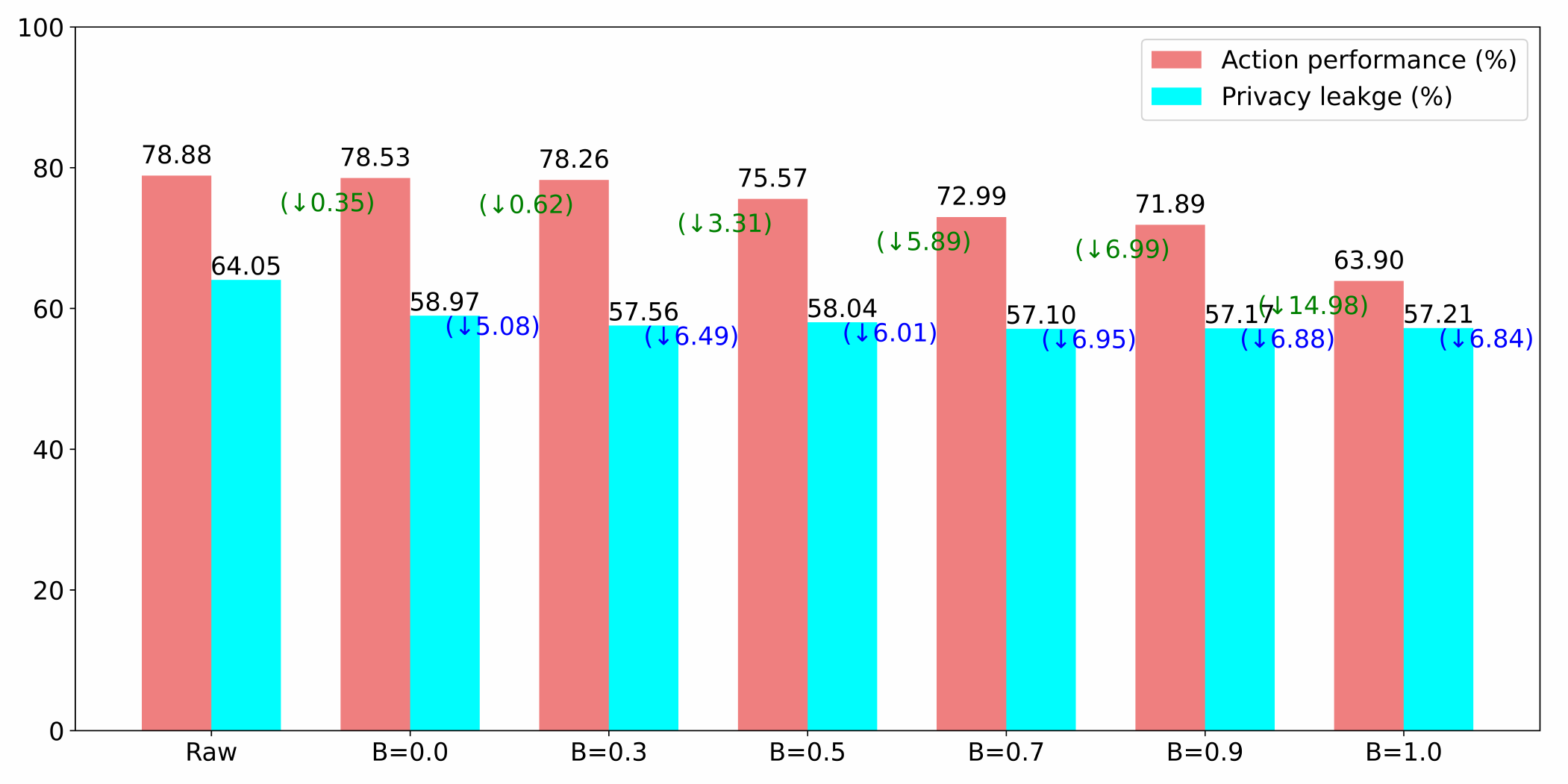}

    \caption{Performance based on different settings of $B$. \textcolor{green}{$\downarrow$} and \textcolor{blue}{$\downarrow$} indicate the performance drop of \textcolor{magenta}{action} and \textcolor{cyan}{privacy} from the raw data respectively.}
    \label{fig:performnace_bar}

\end{figure}

Figure \ref{fig:performnace_bar} depicts the action performance and privacy leakage for different values of $B$ on UCF101-VISPR1. For $B = 0$, no limit is applied on the anonymizer for the utility task, resulting in high action performance but significant privacy leakage. However, $B = 1.0$ indicates complete anonymization, including action features, which severely degrade action performance. Among the different values ($B = 0.3$ to $B = 0.9$), $B = 0.3$ strikes the optimal balance, achieving strong action performance while minimizing privacy leakage, which remains relatively constant from $B = 0.3$ to $B = 0.9$.

% %%%%%%%--------------------------------------------------%%%%%%%%

\subsubsection{Effect of different weights of \texorpdfstring{$\lambda_{\text{penalty}}$}{lambda_penalty}}

The impact of different values of $\lambda_{\text{penalty}}$ on Equation \ref{anonymizer_loss} is presented in Table \ref{tab:lambda_penalty}. This experiment is explicitly conducted for $B = 0.3$, as this value provides the best balance between action performance and privacy preservation in our experiment settings. From Table \ref{tab:lambda_penalty}, we observe that when $\lambda_{\text{penalty}} = 0$, no penalty is applied to the anonymizer, allowing it to obfuscate the input data completely, including essential action features. As a result, action recognition performance declines significantly, reaching levels comparable to \cite{dave2022spact}. However, the cMAP and F1 scores remain relatively unchanged across different values of $\lambda_{\text{penalty}}$, suggesting that this penalty primarily targets action features without impacting the model’s overall privacy leakage.

\begin{table}[ht]
\centering
\resizebox{0.34\textwidth}{!}{
\begin{tabular}{c|c|cc}
\multirow{2}{*}{$\lambda_{\text{penalty}}$} & \textbf{UCF101} & \multicolumn{2}{c}{\textbf{VISPR1}} \\
 & Action  & \multicolumn{2}{c}{Privacy} \\
 & Top-1 ($\uparrow$ \%)  & cMAP ($\downarrow$ \%)  & F1 ($\downarrow$ \%) \\
\hline
0.0  &  63.9      &   57.21     &  0.461     \\
0.1  &  64.8      &   57.51     &  0.461     \\
0.3  &  65.2      &   57.29     &  0.456    \\
0.5  &  71.9      &   57.51     &  0.467    \\
0.7  &  74.28     &   57.60     &  0.466    \\
\textbf{1.0}  &  \textbf{78.26}      &   57.56     &  0.468     
\end{tabular}
}
\caption{Performance comparison of different $\lambda_{\text{penalty}}$ values.}
\label{tab:lambda_penalty}

\end{table}

%%%%%%%--------------------------------------------------%%%%%%%%

%%%%%%%%%%%%%%%%%%%%%%--------------Experiments--------------%%%%%%%%%%%%%%%%%%%%%%

%%%%%%%%%%%%%%%%%%%%%%--------------Conclusion--------------%%%%%%%%%%%%%%%%%%%%%%
\section{Conclusion}
This paper proposes a penalty-driven minimax optimization algorithm that efficiently controls the trade-off between privacy leakage and action performance. We evaluate the proposed model on various penalty settings, such as $B = 0.9, 0.7, 0.5, 0.3$, and compare the performance with other baseline methods. The gradual decline in $B$ led to a significant increase in action performance while privacy leakage remained nearly constant with minimal variation. Our approach demonstrates strong generalization to novel action-private attributes, highlighting the robustness of the proposed method. Experimental results shows that the proposed anonymization function is model agnostic and can be applied to various action classifiers. By incorporating a penalty term into the minimax optimization framework, our framework outperforms existing approaches, which highlight the effectiveness of penalty-driven optimization for privacy-preserving action recognition. Designed for privacy-aware human action recognition tasks, this penalty-driven optimization method can also be leveraged in diverse computer vision applications, which will be further explored in future work.

%%%%%%%%%%%%%%%%%%%%%%--------------Conclusion--------------%%%%%%%%%%%%%%%%%%%%%%

\section*{Acknowledgements}
This work is supported by Milestone Research Program at AAU and Villum Synergy Grant No. 57384.

%%%%%%%%% REFERENCES

 {\small
 \bibliographystyle{ieee_fullname}
 \bibliography{egbib}}

\newpage

\FloatBarrier
\begin{table*}[h!]
	\begin{center}
		\tabcolsep=0.13cm
		\vspace*{-2mm}
		\centering
		\captionsetup{justification=centering}
		\vspace*{-3mm}
		\scalebox{0.7}{
			\begin{tabular}{c}
				\Huge Balancing Privacy and Action Performance: A Penalty-Driven \\ \Huge Approach to Image Anonymization \\ \\ \LARGE -:Supplemetary Material:-
		\end{tabular}}
	\end{center}
	\vspace*{-7mm}
\end{table*}
\FloatBarrier

\newpage

%\begin{refsection}[supp]
The supplementary material is organized into the following sections:
\begin{enumerate}
    \item Section A: Dataset details
    \item Section B: Self supervised contrastive loss for privacy removal branch $f_B$
    \item Section C: Additional Results.
\end{enumerate}

%-------------------------------------------------------------------------

\section{Dataset}
\noindent
\textbf{UCF101 \cite{soomro2012ucf101}} dataset is a large action annotated dataset with 101 different day-to-day human actions with 13,320 videos. All the experiments in this paper are conducted on the split-1, which contains 9,537 training videos and 3,783 testing videos.
\noindent
\\
\textbf{HMDB51 \cite{kuehne2011hmdb}} dataset is comparatively small dataset compared to UCF101 and comprise of 6,849 total videos collected from 51 different human actions. All the results in this papar are reported on split-1, which consists of 3,570 training videos and 1,530 testing videos.
\noindent
\\
\textbf{VISPR \cite{orekondy2017towards}} is a multi-class classification dataset designed for private attribute recognition, comprising 22,167 images annotated with 68 different private attributes, including face, gender, skin color, race, and nudity. Following prior works \cite{dave2022spact, wu2020privacy}, we utilize two distinct subsets of the VISPR dataset, referred as VISPR1 and VISPR2, for our experiments. Each subset contains seven different private attributes, detailed in Table \ref{tab:vispr_attributes}.

\begin{table}[h]
    \centering
    \begin{tabular}{cc}
        \hline
        \textbf{VISPR1}\cite{orekondy2017towards} & \textbf{VISPR2} \cite{orekondy2017towards} \\
        \hline
        $a17\_color$ & $a6\_hair\_color$ \\
        $a4\_gender$ & $a16\_race$ \\
        $a9\_face\_complete$ & $a59\_sports$ \\
        $a10\_face\_partial$ & $a1\_age\_approx$ \\
        $a12\_semi\_nudity$ & $a2\_weight\_approx$ \\
        $a64\_rel\_personal$ & $a73\_landmark$ \\
        $a65\_rel\_soci$ & $a11\_tattoo$ \\
        \hline
    \end{tabular}
    \caption{Private attribute subsets of VISPR\cite{orekondy2017towards} dataset used in experiments.}
    \label{tab:vispr_attributes}
\end{table}

\noindent
\textbf{VPUCF \cite{li2023stprivacy} and VPHMDB \cite{li2023stprivacy}} are large-scale datasets annotated with private attributes for action recognition tasks. The VPUCF dataset is built from the UCF101 dataset, consisting of 101 human action classes with a total of 13,320 videos, while the VPHMDB dataset is derived from the HMDB51 dataset, containing 51 action classes and 6,849 videos. Each video in these datasets is labeled with five private attributes: face, skin color, gender, nudity, and familial relationship. These attributes are represented as binary labels, where 1 indicates the presence of an attribute and 0 denotes its absence. The results reported in this paper are based on experiments conducted on the full dataset.

%------------------------------------------------------------------------

\section{Self supervised contrastive loss for privacy removal branch $f_B$}

\begin{figure*}[t]
    \centering
    \includegraphics[width=\linewidth]{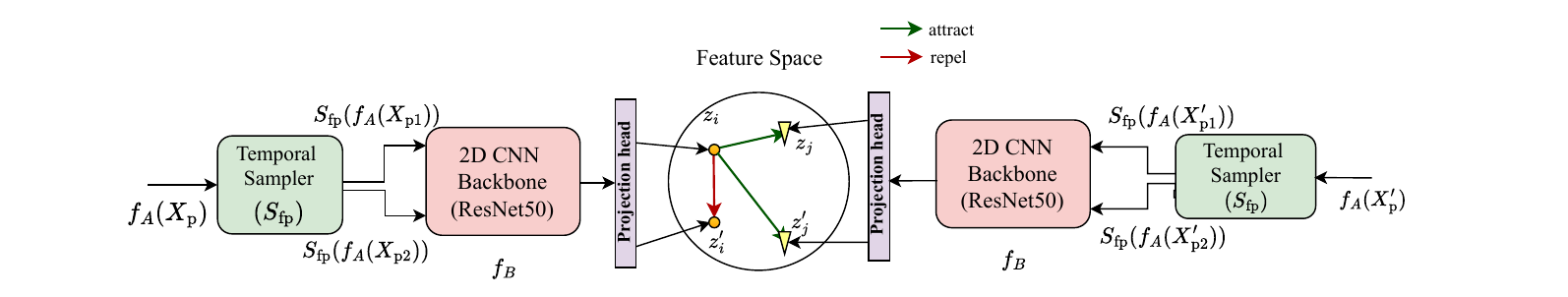}
    \caption{A contrastive learning approach to train the privacy budget task \(f_B\). To anonymize the private attribute of the input data, the distance between the same samples of input data has been maximized, while the distance between the different samples has been minimized.}
    \label{fig:contrastive_loss}
\end{figure*}

A schematic diagram of the self-supervised contrastive loss for the privacy removal branch is depicted in Figure \ref{fig:contrastive_loss}. An input video \( X_p \) is passed through the anonymizer \( f_A \) to generate the anonymized video \( f_A(X_P) \). This anonymized video is then processed by a temporal frame sampler \( S_{\text{fp}}\), which selects two frames based on sampling strategies.  The sampled frame pair \( S_{\text{fp}}(f_A(X_p))\) is then passed through a 2D-CNN backbone, \( f_B \), followed by a non-linear projection head, mapping them into the representation space. This results in two projected representations, \( Z_i \) and \( Z'_i \). The goal of the contrastive loss is to enforce high similarity between projections from the same video \( (Z_i, Z'_i) \) while pushing apart projections from different videos \( (Z_i, Z_j) \) where \( j \neq i \). The NT-Xent contrastive loss \cite{chen2020simple} for a batch of \( N \) videos is formulated as:

\begin{equation}
L^B_i = - \log \frac{h(Z_i, Z'_i)}{\sum_{j=1}^{N} [1[j \neq i] h(Z_i, Z_j) + h(Z_i, Z'_j)]},
\end{equation}

where \( h(u, v) = \exp(\cdot) \) is the similarity function used to compute pairwise relationships in the representation space.

For our anonymization purpose, the contrastive loss function works in the opposite manner compared to \cite{chen2020simple}. Instead of maximizing the agreement between positive pairs and minimizing the agreement between negative pairs, our objective is to increase the disagreement between positive pairs while reducing the agreement between negative pairs. This ensures that the anonymizer struggles to encode private attribute features effectively, thereby enhancing anonymization performance. In the experimental setting, this is achieve by taking the negative gradient. We have selected the positive pairs after every four frames from each video. The rationale is that selecting positive frame pairs from large temporal distances reduces the effectiveness of anonymization. This occurs because incorporating highly dissimilar positive samples in contrastive loss leads to suboptimal representation learning. A similar phenomenon has been reported in prior studies \cite{feichtenhofer2021large, qian2021spatiotemporal}, where using temporally distant positive pairs resulted in degraded performance.

%------------------------------------------------------------------------
\section{Additional Results}

\subsection{Training of $f_A$, $f_B$ and $f_T$}

%%%%%%%--------------Training curve------------%%%%%%%%

\begin{figure*}[h]
    \centering
    
    \subcaptionbox{Training loss curve of anonymizer model $f_A$\label{fig:fa_loss}}
    {\scalebox{0.5}{\includegraphics{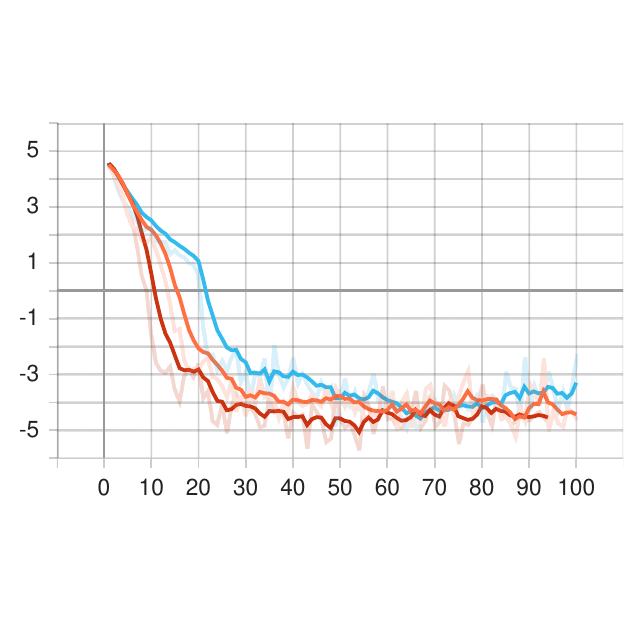}}}
    \hfill
    %\vspace{-5mm}
    \subcaptionbox{Training loss curve of budget task model $f_B$\label{fig:fb_loss}}
     {\scalebox{0.5}{\includegraphics{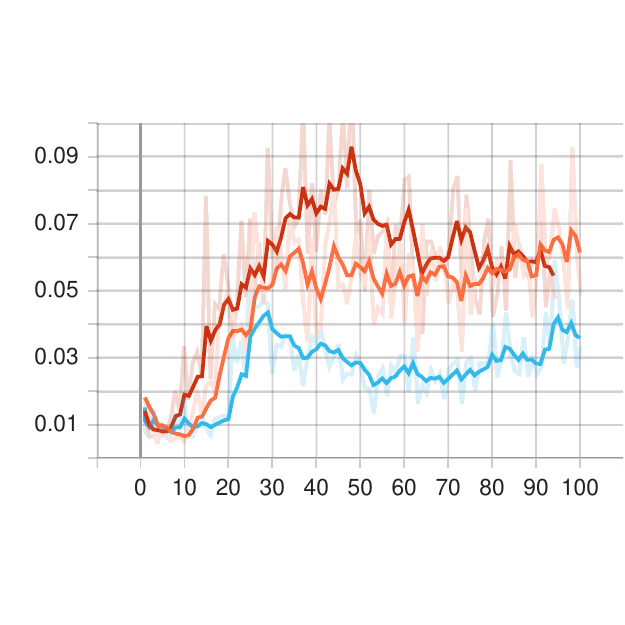}}}
    \hfill
    %\vspace{-5mm}
    \subcaptionbox{Training loss curve of utility task model $f_T$\label{fig:ft_loss}}
    {\scalebox{0.5}{\includegraphics{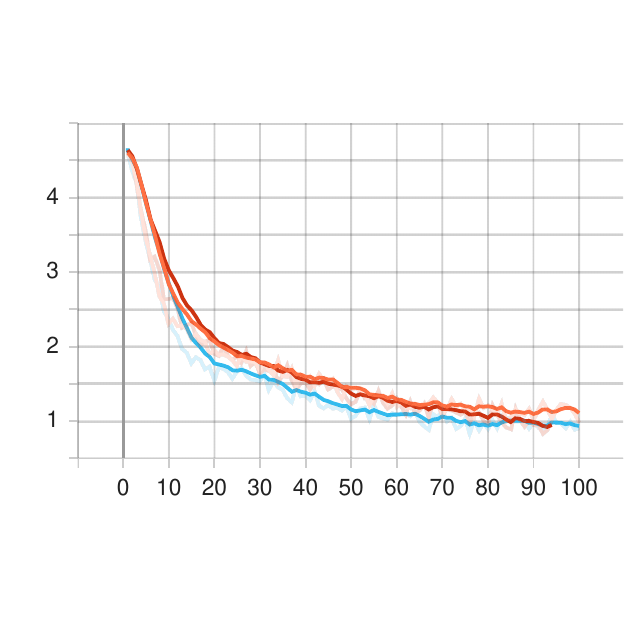}}}
    %\vspace{-2mm}
    \caption{Training loss curves for different functions: (a) Anonymizer $f_A$, (b) Budget task $f_B$, and (c) Utility task $f_T$ for 
    {\textcolor{cyan}{\textbf{$B = 0.5$}}}, 
    {\textcolor{red}{\textbf{$B = 0.7$}}}, and
    {\textcolor{orange}{\textbf{$B = 0.9$}}}.}
    \label{fig:training_loss_curve}
    %\vspace{-4mm}
\end{figure*}

The training loss curves of the anonymizer $f_A$, budget task $f_B$, and utility task $f_T$ are shown in Figure \ref{fig:training_loss_curve}. The anonymizer is expected to converge by minimizing $\mathcal{L_A}$ (refer eq. 3 of main paper), which is reflected in Figure \ref{fig:fa_loss}, where the loss of $f_A$ decreases over several training epochs. Meanwhile, the budget task loss increases, as the anonymizer aims to prevent the encoding of private attribute features of the input data. This trend is observed in Figure \ref{fig:fb_loss}, where the loss of $f_B$ increases with the increase in epochs. In contrast, the utility task loss, which is based on cross-entropy, should decrease as training progresses and eventually converge, as shown in Figure \ref{fig:ft_loss}. Additionally, we observe that incorporating the penalty term $B$ with different values allows $f_A$ to reach convergence while preserving the critical features of the utility task and effectively obstructing the decoding of private attributes in the budget task.

% %%%%%%%--------------------------------------------------%%%%%%%%

\subsection{Evaluate $f_A^*$ on different action classifier $f_T'$}

A learned anonymization function, $f_A^*$, should be able to train any action recognition target model, $f_T'$, on anonymized data without a significant drop in performance. To validate this, we conducted experiments using the learned anonymizer with different utility target models and analyzed the results, as shown in Table \ref{different ft'}. Specifically, we evaluated R3D-18, R2plus1D, MViTv2, and I3D as utility target models. This utility target model is either trained from scratch or initialized with the pretrained weights from the Kinetics 400 dataset. From Table \ref{different ft'}, we observe that with different penalty settings of $B = 0.3, 0.5, 0.7, 0.9$, the performance at $B = 0.3$ is closest to that of raw data. However, as $B$ increases, meaning the level of anonymization is higher, the performance declines. This suggests that the anonymizer effectively anonymizes the incoming data, including action-related features. Notably, when R2plus1D is initialized with pretrained weights from the Kinetics-400 dataset, the action recognition performance improves significantly. This improvement occurs because the model has prior knowledge of action features before training. This experiment also suggested that the proposed anonymization training approach makes the model agnostic, and the learned anonymizer can be used with the different utility target models.

% %%%%%%%------------TABLE-----------%%%%%%%%

\begin{table*}[ht]
\centering
\begin{tabular}{cc|c c c c c c}
\hline
\multicolumn{2}{c|}{\multirow{2}{*}{\textbf{Method}}} & \textbf{R3D-18} & \textbf{R2plus1D} & \begin{tabular}[c]{@{}c@{}}\textbf{R2plus1D}\\ (pretrained on K400)\end{tabular} & \textbf{MViTv2} & \textbf{I3D} & \textbf{C3D} \\
\hline
\multicolumn{2}{c|}{Raw data} & 62.30 & 64.33 & 88.76 & 76.81 & 59.12 & 58.51 \\ 
\multicolumn{2}{c|}{SPACT\cite{dave2022spact}} & 62.03 (\textcolor{cyan}{$\downarrow$ 0.27}) & 62.71 (\textcolor{cyan}{$\downarrow$ 1.62}) &  85.14 (\textcolor{cyan}{$\downarrow$ 3.62}) & -- & -- & 56.10 (\textcolor{cyan}{$\downarrow$ 2.41}) \\ 
\hline
\multirow{4}{*}{Ours} & B = 0.3 & \textbf{62.11} (\textcolor{cyan}{$\downarrow$ 0.19}) & \textbf{63.18} (\textcolor{cyan}{$\downarrow$ 1.15}) & \textbf{86.72} (\textcolor{cyan}{$\downarrow$ 2.04}) & \textbf{73.21} (\textcolor{cyan}{$\downarrow$ 3.6}) & \textbf{58.90} (\textcolor{cyan}{$\downarrow$ 0.22}) & \textbf{57.21} (\textcolor{cyan}{$\downarrow$ 1.3}) 
\\  
& B = 0.5 & 59.01 (\textcolor{cyan}{$\downarrow$ 3.29}) & 60.81 (\textcolor{cyan}{$\downarrow$ 3.52}) & 80.97 (\textcolor{cyan}{$\downarrow$ 7.79}) & 71.04 (\textcolor{cyan}{$\downarrow$ 5.77}) & 56.32 (\textcolor{cyan}{$\downarrow$ 2.8}) & 56.14 (\textcolor{cyan}{$\downarrow$ 2.37}) 
\\  
& B = 0.7 & 57.98 (\textcolor{cyan}{$\downarrow$ 4.32}) & 58.21 (\textcolor{cyan}{$\downarrow$ 6.12}) & 77.12 (\textcolor{cyan}{$\downarrow$ 11.64}) & 69.10 (\textcolor{cyan}{$\downarrow$ 7.71}) & 54.28 (\textcolor{cyan}{$\downarrow$ 4.84}) & 53.81 (\textcolor{cyan}{$\downarrow$ 4.7}) 
\\  
& B = 0.9 & 55.28 (\textcolor{cyan}{$\downarrow$ 7.02}) & 57.92 (\textcolor{cyan}{$\downarrow$ 6.41}) & 76.98 (\textcolor{cyan}{$\downarrow$ 11.78}) & 67.18 (\textcolor{cyan}{$\downarrow$ 9.63}) & 51.11 (\textcolor{cyan}{$\downarrow$ 8.01}) & 50.91 (\textcolor{cyan}{$\downarrow$ 7.6})
\\  
\hline
\end{tabular}
\caption{Comparison of different privacy-preserving methods with different \(f_T'\) architectures trained on UCF101. -- indicates that experiment is not performed on the model. $\downarrow$ indicates drop from the raw data and high value of accuracy considered as better for the action recognition.}
\label{different ft'}
\end{table*}

% %%%%%%%--------------------------------------------------%%%%%%%%

\subsection{Evaluate \texorpdfstring{$f_A^*$}{fA*} on the pretrained \texorpdfstring{$f_b'$}{fB'} on raw data}

%%%%%%-----------------TABLE-----------------%%%%%%%%

\begin{table*}[ht]
\centering
\begin{tabular}{cc|cc|cc|cc}
\hline
\multicolumn{2}{c|}{\multirow{2}{*}{\textbf{Method}}} & \multicolumn{2}{c|}{\textbf{VISPR1}} & \multicolumn{2}{c|}{\textbf{VISPR2}} & \multicolumn{2}{c}{\textbf{PAHMDB}} \\
\multicolumn{2}{c|}{} & cMAP ($\downarrow$ \%) & F1 ($\downarrow$ \%) & cMAP ($\downarrow$ \%) & F1 ($\downarrow$ \%) & cMAP ($\downarrow$ \%) & F1 ($\downarrow$ \%) \\
\hline
\multicolumn{2}{c|}{Raw} & 64.41 & 0.555 & 57.63 & 0.434 & 70.2 & 0.396 \\
\multicolumn{2}{c|}{VITA \cite{wu2020privacy}} & 22.81 (\textcolor{cyan}{$\downarrow$ 41.6}) & 0.243 (\textcolor{cyan}{$\downarrow$ 0.312}) & 26.61 (\textcolor{cyan}{$\downarrow$ 31.02}) & 0.184 (\textcolor{cyan}{$\downarrow$ 0.250}) & 57.01 (\textcolor{cyan}{$\downarrow$ 13.19}) & 0.231 (\textcolor{cyan}{$\downarrow$ 0.165}) \\

\multicolumn{2}{c|}{SPACT \cite{dave2022spact}} & 27.44 (\textcolor{cyan}{$\downarrow$ 36.97}) & 0.076 (\textcolor{cyan}{$\downarrow$ 0.479}) & 20.02 (\textcolor{cyan}{$\downarrow$ 37.61}) & 0.046 (\textcolor{cyan}{$\downarrow$ 0.388}) & 58.90 (\textcolor{cyan}{$\downarrow$ 11.3}) & 0.094 (\textcolor{cyan}{$\downarrow$ 0.165}) \\
\hline
\multirow{4}{*}{Ours} & B = 0.3 & 26.91 (\textcolor{cyan}{$\downarrow$ 37.5}) & 0.081 (\textcolor{cyan}{$\downarrow$ 0.474}) & 20.19 (\textcolor{cyan}{$\downarrow$ 37.44}) & 0.051 (\textcolor{cyan}{$\downarrow$ 0.383}) & 57.19 (\textcolor{cyan}{$\downarrow$ 13.01}) & 0.114 (\textcolor{cyan}{$\downarrow$ 0.165}) \\

& B = 0.5 & 26.24 (\textcolor{cyan}{$\downarrow$ 38.17}) & 0.075 (\textcolor{cyan}{$\downarrow$ 0.480}) & 20.20 (\textcolor{cyan}{$\downarrow$ 37.43}) & 0.052 (\textcolor{cyan}{$\downarrow$ 0.382}) & 57.10 (\textcolor{cyan}{$\downarrow$ 13.10}) & 0.114 (\textcolor{cyan}{$\downarrow$ 0.165}) \\

& B = 0.7 & 26.84 (\textcolor{cyan}{$\downarrow$ 37.57}) & 0.081 (\textcolor{cyan}{$\downarrow$ 0.474}) & 20.15 (\textcolor{cyan}{$\downarrow$ 37.48}) & 0.051 (\textcolor{cyan}{$\downarrow$ 0.383}) & 57.14 (\textcolor{cyan}{$\downarrow$ 13.06}) & 0.112 (\textcolor{cyan}{$\downarrow$ 0.165}) \\

& B = 0.9 & 26.98 (\textcolor{cyan}{$\downarrow$ 37.43}) & 0.079 (\textcolor{cyan}{$\downarrow$ 0.476}) & 20.17 (\textcolor{cyan}{$\downarrow$ 37.46}) & 0.058 (\textcolor{cyan}{$\downarrow$ 0.376}) & 57.13 (\textcolor{cyan}{$\downarrow$ 13.07}) & 0.112 (\textcolor{cyan}{$\downarrow$ 0.165}) \\
\hline
\end{tabular}
\caption{Performance comparison of different methods on privacy leakage evaluation using \textbf{pretrained \(f_B'\)} settings. $\downarrow$ indicates the drop in the performance from the raw data. Lower cMAP and F1 scores indicate better privacy protection. Our method shows almost constant privacy leakage through the different penalty settings and performs better than \cite{dave2022spact}.}
\label{pretrained_fb}
\end{table*}

In a real-world scenario, the trained anonymization model $f_A^*$ is not accessible to anyone. However, there is a potential risk of adversarial attacks targeting the privacy classifier pretrained on raw data, which could lead to the extraction of sensitive privacy-related information. To address this concern, we implemented an additional evaluation protocol. Specifically, we pretrained a new privacy model $f_B'$ (ResNet50) on raw data and subsequently evaluated its performance on anonymized data processed by the learned anonymizer $f_A^*$. The results of this evaluation are presented in Table \ref{pretrained_fb}. Notably, across different penalty setting, $B = 0.3, 0.5, 0.7, 0.9$, the privacy leakage on the dataset remains largely unchanged, with only minor variations. This indicates that incorporating the penalty term in the anonymizer from the utility target model primarily impacts action recognition performance while having minimal influence on privacy leakage. As a result, the anonymizer can effectively anonymize private attributes in the input data to the maximum extent, ensuring minimal privacy leakage through the learned anonymizer. Furthermore, our model demonstrates comparable privacy-preserving performance to \cite{dave2022spact}, while significantly enhancing action recognition performance, as shown in Table \ref{different ft'}.

%%%%%%%--------------------------------------------------%%%%%%%%

\subsection{Effect of different private attribute classifier \texorpdfstring{$f_B'$}{fB'}}

%%%%%%-----------------TABLE-----------------%%%%%%%%

\begin{table*}[ht]
\centering
\resizebox{\textwidth}{!}{
\begin{tabular}{cc|cc|cc|cc|cc|cc}
\hline
\multicolumn{2}{c|}{\multirow{2}{*}{\textbf{Method}}} & \multicolumn{2}{c|}{\textbf{R3D-18}} & \multicolumn{2}{c|}{\textbf{ResNet34}} & \multicolumn{2}{c|}{\textbf{ResNet50}} & \multicolumn{2}{c|}{\textbf{ResNet101}} & \multicolumn{2}{c}{\textbf{ResNet152}}
\\
\multicolumn{2}{c|}{} & cMAP ($\downarrow$ \%) & F1 ($\downarrow$ \%) & cMAP ($\downarrow$ \%) & F1 ($\downarrow$ \%) & cMAP ($\downarrow$ \%) & F1 ($\downarrow$ \%) & cMAP ($\downarrow$ \%) & F1 ($\downarrow$ \%) & cMAP ($\downarrow$ \%) & F1 ($\downarrow$ \%) \\
\hline
\multicolumn{2}{c|}{Raw data} & 64.38 & 0.538 & 65.30 & 0.555 & 64.41 & 0.555 & 60.70 & 0.526 & 58.83 & 0.485 \\

\multicolumn{2}{c|}{SPACT \cite{dave2022spact}} & 54.83 (\textcolor{cyan}{$\downarrow$9.55}) & 0.457 (\textcolor{cyan}{$\downarrow$ 0.081}) & 54.09 (\textcolor{cyan}{$\downarrow$11.21}) & 0.422 (\textcolor{cyan}{$\downarrow$ 0.133}) & 57.43 (\textcolor{cyan}{$\downarrow$ 6.98}) & 0.473 (\textcolor{cyan}{$\downarrow$0.082 }) & 52.94 (\textcolor{cyan}{$\downarrow$ 7.76}) & 0.409 (\textcolor{cyan}{$\downarrow$0.117}) & 53.27 (\textcolor{cyan}{$\downarrow$ 5.56}) & 0.432 (\textcolor{cyan}{$\downarrow$ })
\\ 
\hline
\multirow{4}{*}{Ours} & B = 0.3 & 52.81 (\textcolor{cyan}{$\downarrow$ 11.57}) & 0.431 (\textcolor{cyan}{$\downarrow$ 0.107}) & 52.95 (\textcolor{cyan}{$\downarrow$ 12.35}) & 0.412 (\textcolor{cyan}{$\downarrow$ 0.143}) & 57.41 (\textcolor{cyan}{$\downarrow$ 7.0}) & 0.451 (\textcolor{cyan}{$\downarrow$ 0.104}) & 51.21 (\textcolor{cyan}{$\downarrow$ 9.45}) & 0.391 (\textcolor{cyan}{$\downarrow$0.135}) & 51.25 (\textcolor{cyan}{$\downarrow$7.58}) & 0.422 (\textcolor{cyan}{$\downarrow$ 0.053}) \\  

& B = 0.5 & 52.10 (\textcolor{cyan}{$\downarrow$ 12.28 }) & 0.423 (\textcolor{cyan}{$\downarrow$ 0.115}) & 52.81 (\textcolor{cyan}{$\downarrow$ 12.49 }) & 0.401 (\textcolor{cyan}{$\downarrow$ 0.154}) & 57.32 (\textcolor{cyan}{$\downarrow$ 7.09}) & 0.457 (\textcolor{cyan}{$\downarrow$0.098}) & 51.45 (\textcolor{cyan}{$\downarrow$9.25}) & 0.392 (\textcolor{cyan}{$\downarrow$0.134}) & 51.44 (\textcolor{cyan}{$\downarrow$ 7.39}) & 0.422 (\textcolor{cyan}{$\downarrow$0.063}) \\  

& B = 0.7 & 52.18 (\textcolor{cyan}{$\downarrow$ 12.2}) & 0.429 (\textcolor{cyan}{$\downarrow$ 0.109}) & 52.92 (\textcolor{cyan}{$\downarrow$ 12.38}) & 0.410 (\textcolor{cyan}{$\downarrow$ 0.145 }) & 57.21 (\textcolor{cyan}{$\downarrow$ 7.2}) & 0.452 (\textcolor{cyan}{$\downarrow$ 0.103}) & 51.12 (\textcolor{cyan}{$\downarrow$9.22}) & 0.391 (\textcolor{cyan}{$\downarrow$ 0.135}) & 51.48 (\textcolor{cyan}{$\downarrow$ 7.35}) & 0.421 (\textcolor{cyan}{$\downarrow$ 0.064}) 
\\  

& B = 0.9 & 51.98 (\textcolor{cyan}{$\downarrow$ 12.4}) & 0.435 (\textcolor{cyan}{$\downarrow$ 0.103}) & 52.91 (\textcolor{cyan}{$\downarrow$ 12.39 }) & 0.405 (\textcolor{cyan}{$\downarrow$ 0.150}) & 57.22 (\textcolor{cyan}{$\downarrow$ 7.19}) & 0.452 (\textcolor{cyan}{$\downarrow$ 0.103}) & 51.22 (\textcolor{cyan}{$\downarrow$ 9.48}) & 0.389 (\textcolor{cyan}{$\downarrow$ 0.137}) & 51.51 (\textcolor{cyan}{$\downarrow$ 7.32}) & 0.422 (\textcolor{cyan}{$\downarrow$ 0.063})\\  
\hline
\end{tabular}
}
\caption{Comparison of different privacy-preserving methods with different \(f_B'\) architectures trained on VISPR1. $\downarrow$ indicates the drop in the performance from the raw data. Lower cMAP and F1 scores indicate better privacy protection. Our method shows almost constant privacy leakage through the different penalty settings and performs better than \cite{dave2022spact}.}
\label{different fb'}
\end{table*}

A learned anonymization function $f_A^*$ is designed to protect against privacy leakage from any privacy target model $f_B'$. During the training of the anonymization function, we use ResNet50 as the auxiliary privacy model $f_B$ and evaluate the effectiveness of the learned anonymizer $f_A^*$ on various target privacy classifiers, including R3D-18, R3D-34, R3D-50, R3D-101, and R3D-152, both with and without ImageNet pretraining. As shown in Table \ref{different fb'}, our method effectively prevents privacy leakage, regardless of the chosen target privacy model. Furthermore, across different penalty settings of $B$, the privacy leakage across various target privacy classifiers remains almost constant or only slightly varies. This suggests that introducing the penalty term during the training of the anonymizer does not impact the budget task model $f_B$. Additionally, when ImageNet pretraining is applied, shown in Table \ref{different fb' pretrained}, privacy leakage increases across all methods. However, the relative reduction in leakage compared to the raw data baseline improves, demonstrating the robustness of our approach in mitigating privacy risks.

%%%%%%-----------------TABLE-----------------%%%%%%%%

\begin{table*}[ht]
\centering
\resizebox{\textwidth}{!}{
\begin{tabular}{cc|cc|cc|cc|cc|cc}
\hline
\multicolumn{2}{c|}{\multirow{2}{*}{\textbf{Method}}} & \multicolumn{2}{c|}{\textbf{R3D-18}} & \multicolumn{2}{c|}{\textbf{ResNet34}} & \multicolumn{2}{c|}{\textbf{ResNet50}} & \multicolumn{2}{c|}{\textbf{ResNet101}} & \multicolumn{2}{c}{\textbf{ResNet152}} 
\\
\multicolumn{2}{c|}{} & cMAP ($\downarrow$ \%) & F1 ($\downarrow$ \%) & cMAP ($\downarrow$ \%) & F1 ($\downarrow$ \%) & cMAP ($\downarrow$ \%) & F1 ($\downarrow$ \%) & cMAP ($\downarrow$ \%) & F1 ($\downarrow$ \%) & cMAP ($\downarrow$ \%) & F1 ($\downarrow$ \%) \\
\hline
\multicolumn{2}{c|}{Raw data} & 69.82 & 0.6041 & 69.55 & 0.6447 & 70.76 & 0.6591 & 71.09 & 0.6330 & 69.50 & 0.6130 \\ 
\multicolumn{2}{c|}{SPACT \cite{dave2022spact}} & 59.10 (\textcolor{cyan}{$\downarrow$ 10.72}) & 0.5302 (\textcolor{cyan}{$\downarrow$ 0.0739}) & 59.71 (\textcolor{cyan}{$\downarrow$ 9.84}) & 0.5227 (\textcolor{cyan}{$\downarrow$ 0.122}) & 60.73 (\textcolor{cyan}{$\downarrow$ 10.03}) & 0.5689 (\textcolor{cyan}{$\downarrow$ }0.0902) & 59.24 (\textcolor{cyan}{$\downarrow$ 11.85}) & 0.5601 (\textcolor{cyan}{$\downarrow$ 0.0729}) & 60.51 (\textcolor{cyan}{$\downarrow$ 8.88}) & 0.5352 (\textcolor{cyan}{$\downarrow$ 0.0778})
\\ 
\hline
\multirow{4}{*}{Ours} & B = 0.3 & 58.14 (\textcolor{cyan}{$\downarrow$ 11.68}) & 0.5214 (\textcolor{cyan}{$\downarrow$ 0.0827}) & 57.84 (\textcolor{cyan}{$\downarrow$ 11.71}) & 0.5122 (\textcolor{cyan}{$\downarrow$ 0.1325}) & 58.65 (\textcolor{cyan}{$\downarrow$ 12.11}) & 0.5512 (\textcolor{cyan}{$\downarrow$ 0.1079}) & 58.91 (\textcolor{cyan}{$\downarrow$ 12.18}) & 0.5502 (\textcolor{cyan}{$\downarrow$ 0.0828}) & 59.72 (\textcolor{cyan}{$\downarrow$ 9.78}) & 0.5298 (\textcolor{cyan}{$\downarrow$ 0.0832}) \\  
& B = 0.5 & 58.12 (\textcolor{cyan}{$\downarrow$ 11.7}) & 0.5211 (\textcolor{cyan}{$\downarrow$ 0.083}) & 57.86 (\textcolor{cyan}{$\downarrow$ 11.69}) & 0.5111 (\textcolor{cyan}{$\downarrow$ 0.1336}) & 58.54 (\textcolor{cyan}{$\downarrow$ 12.22}) & 0.5521 (\textcolor{cyan}{$\downarrow$ 0.107}) & 58.95 (\textcolor{cyan}{$\downarrow$ 12.14}) & 0.5509 (\textcolor{cyan}{$\downarrow$ 0.0821}) & 59.85 (\textcolor{cyan}{$\downarrow$ 9.65}) & 0.5296 (\textcolor{cyan}{$\downarrow$ 0.0834}) \\  
& B = 0.7 & 58.21 (\textcolor{cyan}{$\downarrow$ 11.61}) & 0.5212 (\textcolor{cyan}{$\downarrow$ 0.0829}) & 57.91 (\textcolor{cyan}{$\downarrow$ 11.64}) & 0.5214 (\textcolor{cyan}{$\downarrow$ 0.1233}) & 58.98 (\textcolor{cyan}{$\downarrow$ 11.78}) & 0.5525 (\textcolor{cyan}{$\downarrow$ 0.1066}) & 58.72 (\textcolor{cyan}{$\downarrow$ 12.37}) & 0.5519 (\textcolor{cyan}{$\downarrow$ 0.0811}) & 59.96 (\textcolor{cyan}{$\downarrow$ 9.54}) & 0.5284 (\textcolor{cyan}{$\downarrow$ 0.0846}) 
\\  
& B = 0.9 & 58.35 (\textcolor{cyan}{$\downarrow$ 11.47}) & 0.5228 (\textcolor{cyan}{$\downarrow$ 0.0813}) & 57.90 (\textcolor{cyan}{$\downarrow$ 11.65}) & 0.5224 (\textcolor{cyan}{$\downarrow$ 0.1233}) & 58.91 (\textcolor{cyan}{$\downarrow$ 11.85}) & 0.5569 (\textcolor{cyan}{$\downarrow$ 0.1022}) & 58.99 (\textcolor{cyan}{$\downarrow$ 12.10}) & 0.5558 (\textcolor{cyan}{$\downarrow$ 0.0772}) & 59.98 (\textcolor{cyan}{$\downarrow$ 9.52}) & 0.5293 (\textcolor{cyan}{$\downarrow$ 0.0837})
\\  
\hline
\end{tabular}
}
\caption{Comparison of different privacy-preserving methods with different \(f_B'\) architectures trained on VISPR1. The privacy target model is \textbf{pretrained with the ImageNet} weights. $\downarrow$ indicates the drop in the performance from the raw data. Lower cMAP and F1 scores indicate better privacy protection. Our method shows almost constant privacy leakage through the different penalty settings and performs better than \cite{dave2022spact}.}
\label{different fb' pretrained}
\end{table*}

%%%%%%-------------------------------------------------------%%%%%%%%

\subsection{Visualization of anonymized images under different penalty settings of $B$:}
\begin{figure*}[h]
    \centering
    \captionsetup{justification=centering}
    \includegraphics[width=\linewidth]{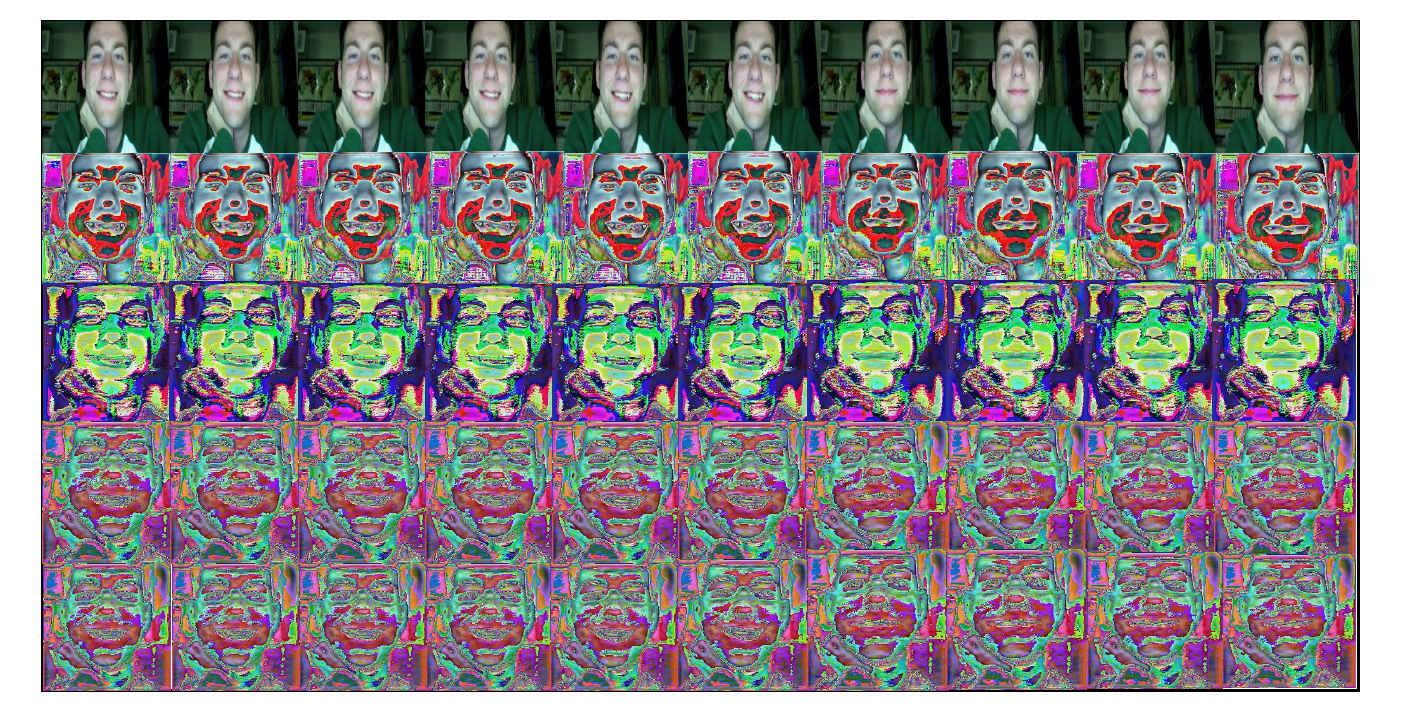}
    \caption{Anonymized frames of smiling action from the HMDB51 dataset across different penalty settings. Top to bottom: Raw image, followed by $B = 0.3$, $B = 0.5$, $B = 0.7$, and $B = 0.9$.}
    \label{fig:smiling}
\end{figure*}

\begin{figure*}[h]
    \centering
    \captionsetup{justification=centering}
    \includegraphics[width=\linewidth]{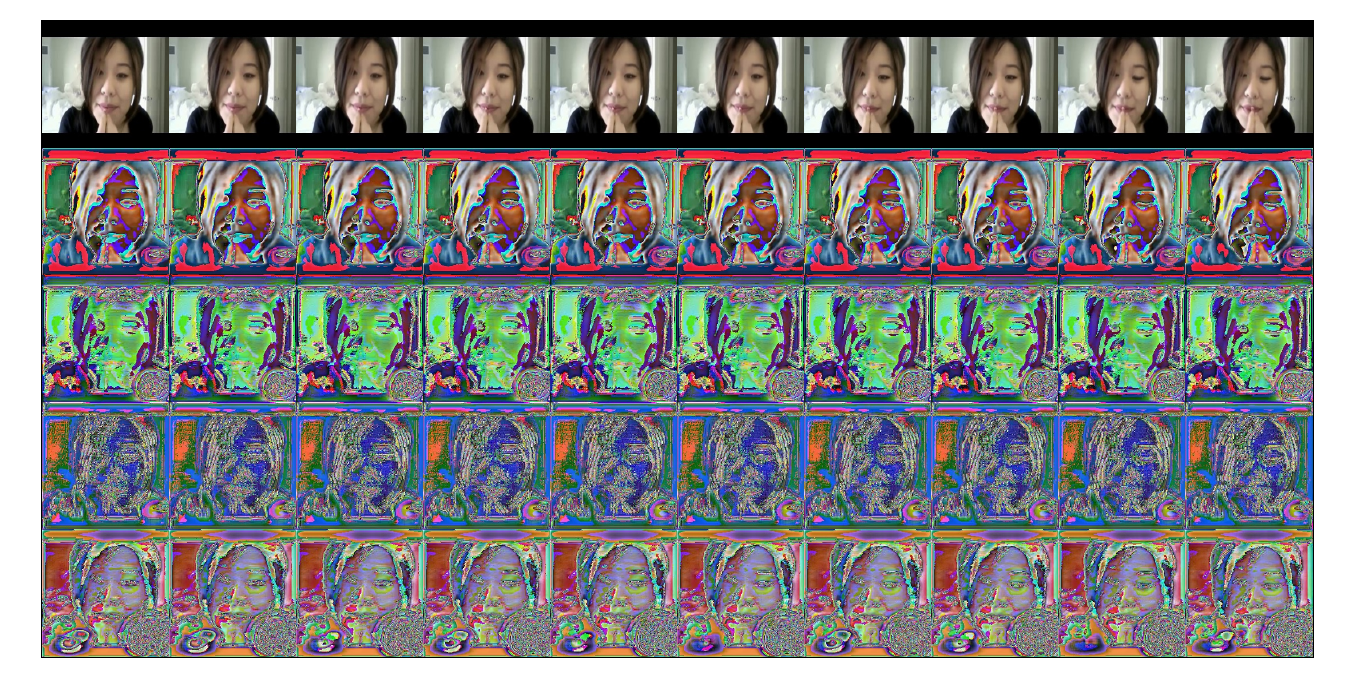}
    \caption{Anonymized frames of apply lipstick action from the UCF101 dataset across different penalty settings. Top to bottom: Raw image, followed by $B = 0.3$, $B = 0.5$, $B = 0.7$, and $B = 0.9$.}
    \label{fig:applylipstick}
\end{figure*}

\begin{figure*}[h]
    \captionsetup{justification=centering}
    \includegraphics[width=\linewidth]{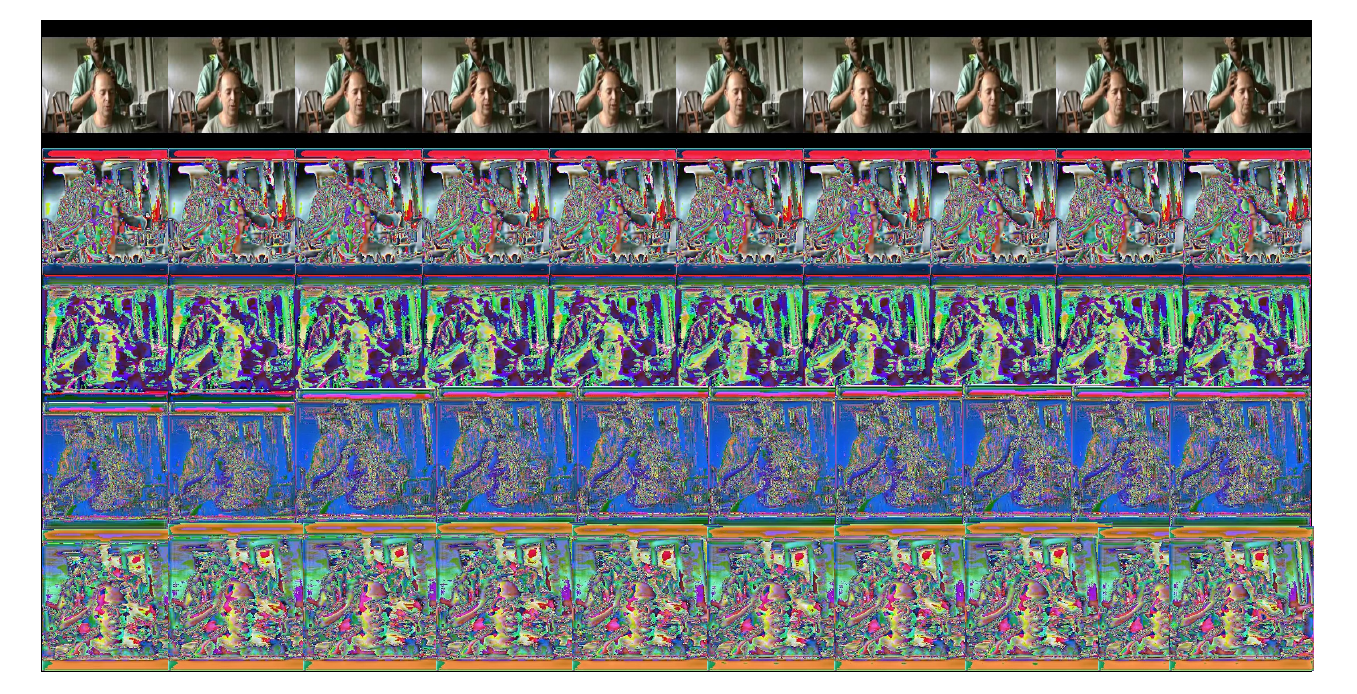}
    \caption{Anonymized frames of head massage action from the UCF101 dataset across different penalty settings. Top to bottom: Raw image, followed by $B = 0.3$, $B = 0.5$, $B = 0.7$, and $B = 0.9$}
    \label{fig:headmassage}
\end{figure*}

To visualize the transformation produced by the learned function $f^*_A$, we present images under different penalty settings of $B$, as shown in the Figure \ref{fig:smiling}, \ref{fig:applylipstick}, \ref{fig:headmassage}. The visualizations indicate that the anonymized images remain primarily consistent across varying penalty values. This is because the penalty is applied explicitly to the action features, while the anonymizer retains complete flexibility to anonymize the images to the maximum extent. Analyzing the Figures \ref{fig:smiling}, \ref{fig:applylipstick}, \ref{fig:headmassage}, we observe that the anonymized images are not identifiable, demonstrating the effective removal of personally identifiable information by our proposed approach.

\end{document}